\documentclass[sigconf, screen, nonacm]{acmart}

\usepackage{booktabs}
\usepackage{multirow}
\usepackage{amsmath}
\usepackage{mathtools}
\usepackage{graphicx}
\usepackage{float}
\usepackage{subcaption}
\usepackage{algorithm}
\usepackage{algpseudocode}
\usepackage[table,xcdraw]{xcolor}
\usepackage{enumitem}
\usepackage{array}
\usepackage{url}

\AtBeginDocument{%
  }

\setcopyright{none}
\renewcommand\footnotetextcopyrightpermission[1]{}

\title{Quality-Aware Robust Multi-View Clustering for Heterogeneous Observation Noise}

\author{Peihan Wu}
\email{22321313@zju.edu.cn}
\affiliation{%
  \institution{Zhejiang University}
  \city{Hangzhou}
  \country{China}}

\author{Guanjie Cheng}
\authornote{Corresponding author.}
\email{chengguanjie@zju.edu.cn}
\affiliation{%
  \institution{Zhejiang University}
  \city{Hangzhou}
  \country{China}}

\author{Yufei Tong}
\email{yufeitong@zju.edu.cn}
\affiliation{%
  \institution{Zhejiang University}
  \city{Hangzhou}
  \country{China}}

\author{Meng Xi}
\email{ximeng@zju.edu.cn}
\affiliation{%
  \institution{Zhejiang University}
  \city{Hangzhou}
  \country{China}}

\author{Shuiguang Deng}
\email{dengsg@zju.edu.cn}
\affiliation{%
  \institution{Zhejiang University}
  \city{Hangzhou}
  \country{China}}

\renewcommand{\shortauthors}{Wu et al.}

\begin{document}

\begin{abstract}
Deep multi-view clustering has achieved remarkable progress but remains vulnerable to complex noise in real-world applications. Existing noisy robust methods predominantly rely on a simplified binary assumption, treating data as either perfectly clean or completely corrupted. This overlooks the prevalent existence of heterogeneous observation noise, where contamination intensity varies continuously across data. To bridge this gap, we propose a novel framework termed Quality-Aware Robust Multi-View Clustering (QARMVC). Specifically, QARMVC employs an information bottleneck mechanism to extract intrinsic semantics for view reconstruction. Leveraging the insight that noise disrupts semantic integrity and impedes reconstruction, we utilize the resulting reconstruction discrepancy to precisely quantify fine-grained contamination intensity and derive instance-level quality scores. These scores are integrated into a hierarchical learning strategy: at the feature level, a quality-weighted contrastive objective is designed to adaptively suppress the propagation of noise; at the fusion level, a high-quality global consensus is constructed via quality-weighted aggregation, which is subsequently utilized to align and rectify local views via mutual information maximization. Extensive experiments on five benchmark datasets demonstrate that QARMVC consistently outperforms state-of-the-art baselines, particularly in scenarios with heterogeneous noise intensities.
\end{abstract}

\ccsdesc[500]{Computing methodologies~Unsupervised learning}
\ccsdesc[300]{Computing methodologies~Neural networks}
\ccsdesc[300]{Information systems~Multimedia and multimodal retrieval}

\keywords{multi-view clustering, information bottleneck, contrastive learning, mutual information}

\maketitle

\section{Introduction}

In recent years, with the rapid advancement of sensing, communication, and data acquisition technologies, multi-view data has become increasingly prevalent in a wide range of real-world applications, such as industrial anomaly detection, social recommendation~\cite{chen2024multi}, and urban health profiling and prediction~\cite{li2025curegraph}. Unlike single-view data, multi-view data describes the same object from multiple heterogeneous perspectives, where different views may capture distinct yet complementary characteristics. Such a multi-perspective description provides richer semantic cues and stronger structural information, making it possible to uncover latent patterns that are difficult to identify from any individual view alone. As a result, effectively exploiting the consistency and complementarity across multiple views has become a central issue in unsupervised representation learning and data mining.

\begin{figure}[t]
\centering
\includegraphics[scale=0.3]{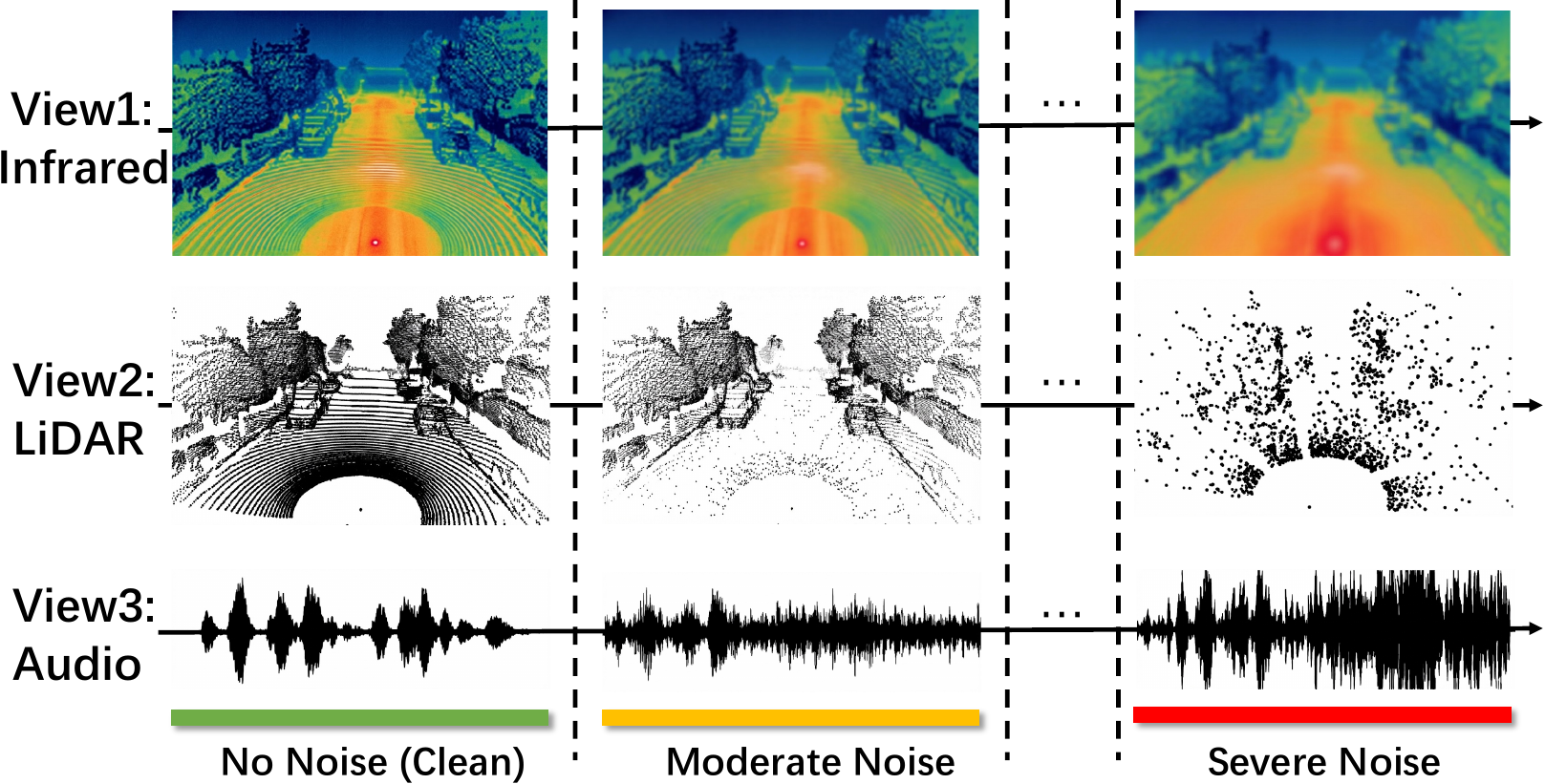}
\caption{An illustrative diagram of heterogeneous noise intensity in a multi-view scenario. The diagram displays Infrared, LiDAR, and Audio views under varying environmental conditions. From left to right, the data quality exhibits a continuous degradation process from clean to severe noise, rather than a simple binary state.}
\label{motivation}
\end{figure}

To address this problem, Multi-View Clustering (MVC) has been widely studied as a fundamental unsupervised task, aiming to partition unlabeled data by jointly leveraging information from multiple views. Early MVC approaches were mainly built upon traditional machine learning paradigms, including multi-view subspace learning~\cite{DBLP:conf/aaai/KangZZSHX20,gao2015multi}, non-negative matrix factorization based methods~\cite{liu2013multi,huang2020auto}, and graph-based clustering frameworks~\cite{suyuan_AAAI,tang2020cgd,wang2019gmc}. These methods promote cross-view information fusion by constructing shared latent spaces, enforcing consistency constraints, or learning consensus graph structures. Although they have achieved encouraging results, their representation capacity is often limited by shallow architectures, hand-crafted similarity modeling, or relatively restrictive assumptions on data distributions. Benefiting from the strong nonlinear modeling ability of deep neural networks, Deep Multi-View Clustering (DMVC) has gradually become the dominant paradigm in this field~\cite{trosten2021reconsidering}~\cite{xu2023selfsupervised}~\cite{yang2023dealmvc}~\cite{chen2023cvcl}~\cite{cui2024dcmvc}. Representative methods incorporate adversarial learning~\cite{li2019deep}, contrastive feature learning~\cite{xu2022multi}~\cite{wang2024sce}~\cite{wang2024viewgap}~\cite{cai2024mcpl}, and generative modeling~\cite{wen2024diffusion} into multi-view clustering, substantially improving the quality of learned representations and the discriminability of clustering structures.

Despite the significant progress of existing DMVC methods, their robustness is still far from satisfactory when confronted with noisy real-world environments. Most existing studies model observation noise under a simplified binary assumption~\cite{MVCAN,yang2025automatically}, where each sample or view is regarded as either entirely clean or completely corrupted. However, such a coarse-grained assumption rarely holds in practice. Real-world multi-view data is more likely to suffer from heterogeneous observation noise, in which different instance-view pairs exhibit different and continuously varying levels of contamination~\cite{dong2025rac}. As shown in Figure~1, sensory quality does not evolve in a discrete clean-versus-noisy manner, but instead spans a continuous spectrum from high-fidelity observations to mild degradation and even severe corruption. For example, in autonomous driving systems involving Infrared, LiDAR, and audio modalities, environmental disturbances such as adverse weather, motion blur, occlusion, and signal interference may affect different views to different extents. Under such conditions, some observations may still retain substantial semantic information despite moderate corruption, whereas others may become highly unreliable. This heterogeneous contamination pattern makes robust cross-view fusion substantially more difficult. Unfortunately, existing methods usually cannot explicitly characterize the fine-grained contamination intensity of each observation. Simply discarding non-ideal views as outliers may lead to the loss of useful complementary semantics, while blindly aggregating all views may introduce unreliable information into the common semantic space and deteriorate clustering quality. Therefore, how to accurately estimate the contamination intensity of each instance-view pair and conduct effective semantic learning under varying noise levels remains an urgent and largely underexplored problem in multi-view clustering.

To bridge this gap, we propose a \textbf{Q}uality-\textbf{A}ware \textbf{R}obust \textbf{M}ulti-\textbf{V}iew \textbf{C}lustering framework, termed \textbf{QARMVC}. Specifically, we employ the information bottleneck mechanism to capture intrinsic semantics by compressing each view into a compact latent space. Based on the insight that noise disrupts semantic integrity and impedes recovery, the resulting reconstruction discrepancy is utilized to precisely quantify heterogeneous contamination intensity, yielding fine-grained, instance-level quality scores. On this basis, we implement a hierarchical learning strategy. At the feature level, we utilize autoencoder-based embedding combined with a quality-weighted contrastive objective to ensure the semantic stability of the latent space. At the fusion level, view-specific embeddings are aggregated via quality-weighted fusion to construct a robust global consensus. We then maximize mutual information between this global target and local representations, effectively guiding noisy views to recover consistent semantics. 

The key contributions of our paper are summarized as follows:
\begin{itemize}[leftmargin=1.5em]
\item We propose a novel Quality-Aware Robust Multi-View Clustering framework (QARMVC) to function robustly in heterogeneous noisy environments. To the best of our knowledge, this work is the first to incorporate fine-grained noise scores into multi-view clustering in a systematic manner.
\item To perceive the heterogeneous contamination intensities, we introduce an information bottleneck mechanism to precisely quantify the quality of data. We further design a quality-weighted contrastive loss to ensure feature stability and construct a high-quality global representation to guide the rectification of contaminated views via mutual information maximization, thereby effectively alleviating noise interference.
\item Extensive experiments on five benchmark datasets demonstrate that QARMVC consistently outperforms state-of-the-art baselines in both clustering accuracy and robustness under varying noise intensities.
\end{itemize}

\section{Related Work}
\subsection{Deep Multi-View Clustering}

Deep multi-view clustering (DMVC) has attracted increasing attention due to the strong representation learning capability of deep neural networks. Compared with traditional multi-view clustering methods based on subspace learning~\cite{gao2015multi,DBLP:conf/aaai/KangZZSHX20}, matrix factorization~\cite{liu2013multi,huang2020auto}, and graph fusion~\cite{tang2020cgd,wang2019gmc}, DMVC is able to learn nonlinear view-specific representations and cross-view interactions in an end-to-end manner, thereby showing stronger adaptability to complex data distributions. Existing DMVC methods can be roughly grouped into several representative lines. The first line follows clustering-oriented deep embedding, where feature learning and cluster structure optimization are jointly conducted in a unified framework. For example, DAMC~\cite{li2019deep} introduces adversarial learning to enhance the consistency of latent representations across views. The second line emphasizes contrastive representation learning, which improves clustering quality by strengthening agreement between semantically related samples or views. Representative methods such as~\cite{xu2022multi,DIVIDE} exploit multi-level or decoupled contrastive objectives to enhance the discriminability and robustness of learned features. The third line focuses on graph-structured multi-view modeling, where relational information is explicitly encoded to preserve local topology and facilitate cross-view fusion~\cite{tang2020cgd,wang2019gmc}. In addition, recent studies have further explored generative modeling for multi-view representation learning, such as diffusion-based frameworks that improve semantic completion and feature consistency under complex scenarios~\cite{wen2024diffusion}.

\subsection{Noisy Multi-View Clustering}

Despite the success of DMVC in handling multi-modal data, its performance often degrades in real-world applications due to complex noise. Generally, existing robust MVC research categorizes these challenges into three distinct scenarios: correspondence noise, missing noise, and absolute observation noise. Specifically, regarding correspondence noise, characterized by mismatches between samples, representative works~\cite{sun2024robust}~\cite{sun2025roll}~\cite{huang2020pvc}~\cite{he2024vital} tackle this misalignment by designing noise-tolerant contrastive losses or leveraging inter-view similarity contexts to rectify erroneous alignments. Missing noise deals with partial data loss, where imputation-based methods~\cite{tang2022deep}~\cite{yang2022robust}~\cite{jin2023deep} utilize observable neighboring samples to recover information, while non-imputation methods~\cite{xu2022deep}~\cite{DIVIDE}~\cite{zhang2025incomplete} rely on cross-view prediction in semantic space. Finally, absolute observation noise typically operates under a binary assumption, treating views as either entirely clean or completely corrupted~\cite{MVCAN}. Proactive methods address this by identifying and handling outliers. Specifically, AIRMVC~\cite{yang2025automatically} formulates this as an anomaly detection problem to distinguish clean samples from noisy outliers. RAC-DMVC~\cite{dong2025rac} employs a cross-view cross-reconstruction mechanism, leveraging the information from reliable views to rectify the corrupted ones.

While the aforementioned studies have made significant strides, they overlook a more prevalent and realistic scenario: heterogeneous observation noise~\cite{wang2023multimodal}. Existing methods for absolute noise have limited efficacy here, as they typically rely on at least one clean view per instance to identify and rectify noisy counterparts. However, in heterogeneous scenarios where data represents a mixture of valid semantics and noise, this binary treatment is ineffective: simply regarding such data as outliers overlooks the intrinsic semantic information, while indiscriminately fusing them leads to the distortion of the common semantic space. Therefore, bridging this gap requires a quality-aware framework capable of perceiving fine-grained noise intensities to balance semantic retention and noise suppression.

\begin{figure*}[t]
    \centering
    \includegraphics[width=0.90\linewidth]{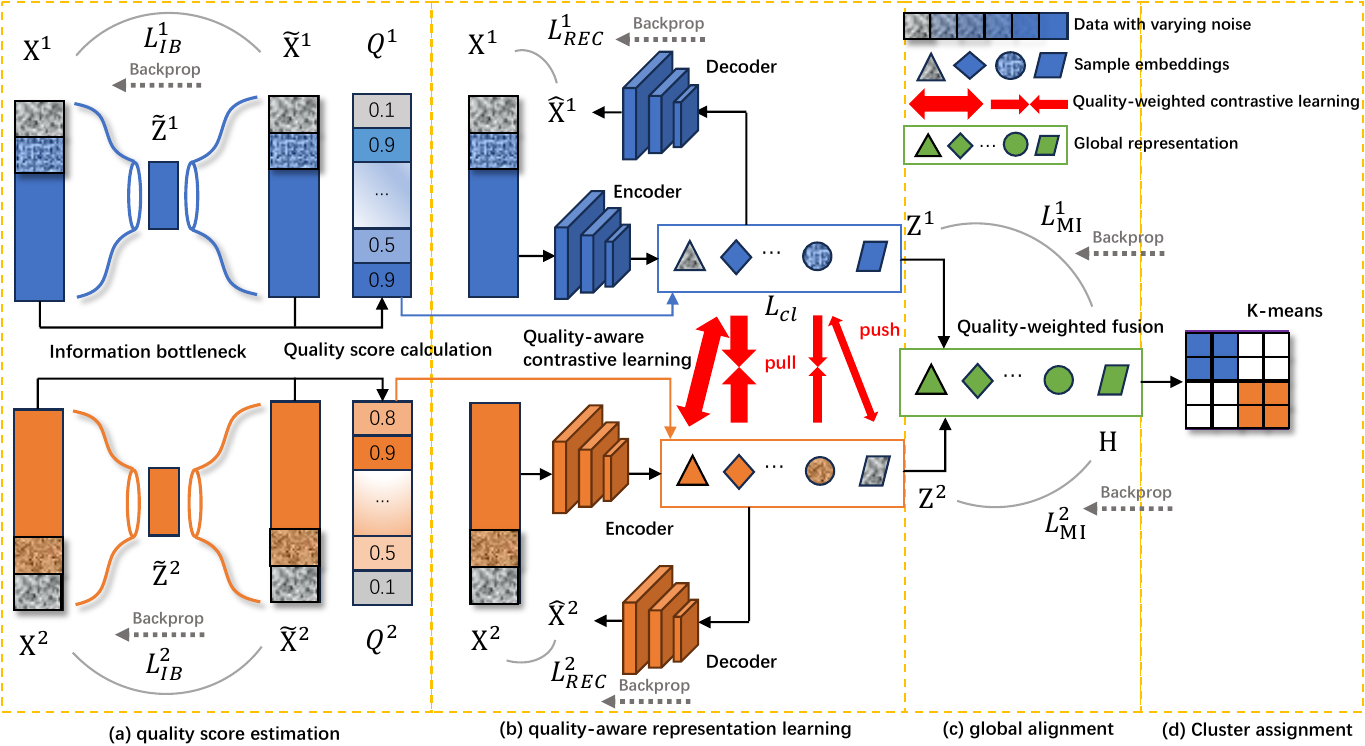}
    \caption{
        The framework consists of four modules: (a) Quality score estimation, where an information bottleneck mechanism quantifies heterogeneous noise intensity to derive instance-view specific quality scores; (b) Quality-aware representation learning, which utilizes these scores to re-weight contrastive learning, thereby suppressing noisy anchors; (c) Global alignment, where local views are aligned with a robust high-quality global consensus via mutual information maximization; (d) Cluster assignment: Perform K-means clustering on the global consensus representation $H$ to obtain the final cluster labels.
    }
    \label{overall}
\end{figure*}

\section{Methodology}

\subsection{Preliminary}

Given a multi-view dataset $\mathcal{X} = \{X^1, X^2, \dots, X^V\}$ consisting of $N$ instances across $V$ views, the data matrix for the $v$-th view is denoted as $X^v = [\mathbf{x}_1^v, \mathbf{x}_2^v, \dots, \mathbf{x}_N^v] \in \mathbb{R}^{d_v \times N}$, where $\mathbf{x}_i^v \in \mathbb{R}^{d_v}$ represents the feature vector of the $i$-th instance in the $v$-th view. In real-world scenarios, the observed data typically suffers from heterogeneous observation noise, meaning that the dataset $\mathcal{X}$ consists of samples with non-uniform quality and varying levels of degradation. The primary objective is to partition the unlabeled dataset $\mathcal{X}$ into $K$ disjoint clusters.

\subsection{Quality Score Estimation}

High-dimensional observations usually exhibit a low-dimensional intrinsic structure, enabling the extraction of core semantics by compressing the raw input into a compact latent variable. To derive a semantically rich representation, we construct an information bottleneck mechanism~\cite{alemi2017deep}. Our objective is to maximize the mutual information between the input view and its latent representation while constraining the capacity of the latent space:
\begin{equation}
\max I(X^v; \tilde{Z}) \quad \text{s.t.} \quad \dim(\tilde{Z}) \ll \dim(X^v).
\end{equation}
Here, maximizing $I(X^v; \tilde{Z})$ ensures the preservation of intrinsic information from the original view $X^v$, while the dimensionality constraint prevents the latent variable from learning a trivial identity mapping. From an information-theoretic perspective, we first expand the mutual information term:
\begin{equation}
\begin{split}
I(X^v; \tilde{Z})
&= \iint p(\mathbf{x}^v, \tilde{\mathbf{z}}) \log \frac{p(\mathbf{x}^v|\tilde{\mathbf{z}})}{p(\mathbf{x}^v)} \,d\mathbf{x}^v\,d\tilde{\mathbf{z}} \\
&= \iint p(\mathbf{x}^v) p(\tilde{\mathbf{z}}|\mathbf{x}^v) \log p(\mathbf{x}^v|\tilde{\mathbf{z}}) \,d\mathbf{x}^v\,d\tilde{\mathbf{z}} + H(X^v) \\
&\ge \iint p(\mathbf{x}^v) p(\tilde{\mathbf{z}}|\mathbf{x}^v) \log p(\mathbf{x}^v|\tilde{\mathbf{z}}) \,d\mathbf{x}^v\,d\tilde{\mathbf{z}},
\end{split}
\label{eq:mi_first_bound}
\end{equation}
where $\mathbf{x}^v$ and $\tilde{\mathbf{z}}$ are instances of $X^v$ and $\tilde{Z}$, respectively. Since the true conditional distribution $p(\mathbf{x}^v|\tilde{\mathbf{z}})$ is unknown, we approximate it using a stochastic decoder $q^v(\mathbf{x}^v|\tilde{\mathbf{z}})$, which generates the reconstructed data $\tilde{\mathbf{x}}^v$. We further decompose the integral in Eq.~\eqref{eq:mi_first_bound} to derive a tractable variational lower bound:
\begin{equation}
\begin{split}
I(X^v; \tilde{Z})
&\ge \iint p(\mathbf{x}^v) p(\tilde{\mathbf{z}}|\mathbf{x}^v) \log q^v(\mathbf{x}^v|\tilde{\mathbf{z}}) \,d\mathbf{x}^v\,d\tilde{\mathbf{z}} \\
&\quad + \iint p(\mathbf{x}^v) p(\tilde{\mathbf{z}}|\mathbf{x}^v) \log \frac{p(\mathbf{x}^v|\tilde{\mathbf{z}})}{q^v(\mathbf{x}^v|\tilde{\mathbf{z}})} \,d\mathbf{x}^v\,d\tilde{\mathbf{z}} \\
&\ge \mathbb{E}_{\mathbf{x}^v \sim p(\mathbf{x}^v)} \Big[ \int p(\tilde{\mathbf{z}}|\mathbf{x}^v) \log q^v(\mathbf{x}^v|\tilde{\mathbf{z}}) \,d\tilde{\mathbf{z}} \Big].
\end{split}
\label{eq:mi_second_bound}
\end{equation}
By approximating the latent distribution $p(\tilde{\mathbf{z}}|\mathbf{x}^v)$ via a stochastic encoder, maximizing this lower bound is equivalent to minimizing the negative log-likelihood loss:
\begin{equation}
\mathcal{L}_{IB}^v = - \mathbb{E}_{\mathbf{x}^v \sim p(\mathbf{x}^v)} \Big[ \mathbb{E}_{\tilde{\mathbf{z}} \sim p(\tilde{\mathbf{z}}|\mathbf{x}^v)} \big[ \log q^v(\mathbf{x}^v|\tilde{\mathbf{z}}) \big] \Big].
\label{eq:ib_loss}
\end{equation}

After training, clean samples adhering to the intrinsic manifold can be accurately reconstructed. Conversely, contaminated samples, disrupted by stochastic noise, fail to yield valid latent representations through the compression bottleneck, leading to significant reconstruction discrepancies. We quantify this contamination using the instance-level reconstruction error $R_i^v = \lVert \mathbf{x}_i^v - \tilde{\mathbf{x}}_i^v \rVert_1$. To facilitate view-adaptive weighting, we define a normalized contamination score:
\begin{equation}
C_i^v = \frac{R_i^v - \min(R^v)}{\max(R^v) - \min(R^v)},
\label{eq:contamination_score}
\end{equation}
where $\max(R^v)$ and $\min(R^v)$ denote the extreme reconstruction errors computed across all samples in view $v$. The final quality score is obtained as $Q_i^v = (1 - C_i^v)^2$. This score acts as a dynamic weighting factor in the subsequent joint training phase, adaptively suppressing the influence of low-quality data.

\subsection{Multi-view Representation Learning}

With the estimated quality scores, we employ independent deep autoencoders to extract clustering-friendly features for each view. For the $v$-th view, the encoder $\mathcal{E}^v$ maps the input $\mathbf{x}_i^v$ into a latent representation $\mathbf{z}_i^v = \mathcal{E}^v(\mathbf{x}_i^v)$, while the decoder $\mathcal{D}^v$ generates the reconstruction $\hat{\mathbf{x}}_i^v = \mathcal{D}^v(\mathbf{z}_i^v)$. To ensure effective feature extraction, the model minimizes the aggregate reconstruction loss:
\begin{equation}
\mathcal{L}_{REC} = \sum_{v=1}^V \sum_{i=1}^N \lVert \mathbf{x}_i^v - \hat{\mathbf{x}}_i^v \rVert_2^2.
\label{loss_rec}
\end{equation}

Building upon these latent representations, we introduce a contrastive learning mechanism to mitigate cross-view heterogeneity and capture consistent semantics. The semantic similarity between representations is quantified using cosine similarity:
\begin{equation}
S(\mathbf{z}_i^u, \mathbf{z}_j^v) = \frac{(\mathbf{z}_i^u)^\top \mathbf{z}_j^v}{\lVert \mathbf{z}_i^u \rVert_2 \lVert \mathbf{z}_j^v \rVert_2}.
\label{similarity}
\end{equation}
To align information across perspectives, we formulate an instance-level contrastive objective. Specifically, taking $\mathbf{z}_i^u$ from view $u$ as the anchor and view $v$ ($u \neq v$) as the target, the contrastive loss for the $i$-th instance is defined as:
\begin{equation}
\ell_{i}^{(u \rightarrow v)} = - \log \frac{\exp(S(\mathbf{z}_i^u, \mathbf{z}_i^v) / \tau)}{\sum_{k=1}^N \exp(S(\mathbf{z}_i^u, \mathbf{z}_k^v) / \tau)},
\label{loss_instance}
\end{equation}
where $\tau$ denotes the temperature hyperparameter. Minimizing Eq.~\eqref{loss_instance} effectively pulls the positive counterpart $\mathbf{z}_i^v$ toward the anchor $\mathbf{z}_i^u$ in the latent space, while simultaneously pushing the negative samples $\{\mathbf{z}_k^v\}_{k \neq i}$ away.

Standard contrastive frameworks typically treat all anchors equally by minimizing the average of $\ell_{i}^{(u \rightarrow v)}$. However, such an indiscriminate strategy is suboptimal in the presence of heterogeneous observation noise; aligning corrupted anchors with other views inevitably propagates noise and distorts the common semantic space. To address this, we propose a quality-aware robust contrastive loss by incorporating the instance-view quality score $Q_i^u$. By adaptively re-weighting the contribution of each anchor based on its reliability, the final objective is formulated as:
\begin{equation}
\mathcal{L}_{RCL} = \sum_{i=1}^N \sum_{u=1}^V \sum_{v \neq u} Q_i^u \cdot \ell_{i}^{(u \rightarrow v)}.
\label{loss_con}
\end{equation}
This formulation ensures that high-quality instances dominate the semantic alignment process, while the negative impact of contaminated data is suppressed, facilitating the learning of robust, view-consistent representations.

\subsection{Quality-Guided Global Fusion and Alignment}

To leverage cross-view complementarity while maintaining robustness, we introduce a global-local alignment module consisting of quality-guided fusion and mutual information (MI) maximization~\cite{lin2022dual}. Standard fusion strategies~\cite{ding2025incomplete}, such as simple concatenation or averaging, treat all views indiscriminately, which often leads to the contamination of the common semantic space by noisy observations. To address this, we utilize the estimated quality scores to construct a robust global representation.

Specifically, we first normalize the instance-view quality scores to obtain the fusion weights $w_i^v = Q_i^v / \sum_{k=1}^V Q_i^k$. Subsequently, the global consensus representation $\mathbf{h}_i$ for the $i$-th instance is generated via a weighted aggregation of view-specific embeddings:
\begin{equation}
\mathbf{h}_i = \sum_{v=1}^V w_i^v \mathbf{z}_i^v,
\label{eq_global_fusion}
\end{equation}
where $\mathbf{z}_i^v$ denotes the embedding of the $i$-th instance in the $v$-th view.
By prioritizing high-quality views via Eq.~\eqref{eq_global_fusion}, the global representation $H$ successfully extracts representative semantics while adaptively suppressing noise interference.

To further utilize this robust global consensus to guide and rectify view-specific learning, we maximize the consistency between $H$ and each local representation $Z^v$. This objective is formulated as minimizing the negative mutual information loss:
\begin{equation}
\begin{split}
\mathcal{L}_{MI} &= - \sum_{v=1}^V I(H; Z^v) \\
&= - \sum_{v=1}^V \sum_{i,j} p(\mathbf{h}_i, \mathbf{z}_j^v) \log \frac{p(\mathbf{h}_i, \mathbf{z}_j^v)}{p(\mathbf{h}_i) p(\mathbf{z}_j^v)},
\end{split}
\label{eq_mi_loss}
\end{equation}
where $p(\mathbf{h}_i, \mathbf{z}_j^v)$ denotes the joint probability distribution, and $p(\mathbf{h}_i)$, $ p(\mathbf{z}_j^v)$ are the corresponding marginals.

Minimizing $\mathcal{L}_{MI}$ aligns view-specific features with the global consensus. This high-quality target provides reliable semantic guidance to rectify local inconsistencies and distortions in contaminated views, thereby fostering a unified and robust latent space on which the final clustering is performed.

\subsection{Overall Algorithm}

The comprehensive objective function of QARMVC integrates quality-weighted reconstruction, robust contrastive learning and mutual information alignment. The total loss is formulated as:
\begin{equation}
\mathcal{L} = \lambda_1 \mathcal{L}_{REC} + \lambda_2 \mathcal{L}_{RCL} + \lambda_3 \mathcal{L}_{MI},
\label{loss_total_final}
\end{equation}
where $\lambda_1$, $\lambda_2$ and $\lambda_3$ are trade-off hyperparameters. The optimization process, detailed in Algorithm~\ref{alg:qarmvc}.

\begin{algorithm}[t]
   \caption{Training Procedure of QARMVC}
   \label{alg:qarmvc}
\begin{algorithmic}[1]
   \Require Multi-view dataset $\mathcal{X}=\{X^v\}_{v=1}^{V}$, cluster number $K$, max epochs $E$
   \Ensure Clustering results $\mathbf{y}$
   \State Train the information bottleneck module by minimizing Eq.~\eqref{eq:ib_loss};
   \State Compute the contamination scores via Eq.~\eqref{eq:contamination_score} and obtain the quality scores $Q_i^v=(1-C_i^v)^2$;
   \For{$epoch = 1$ \textbf{to} $E$}
       \State Encode each view to obtain latent representations $\{Z^v\}_{v=1}^{V}$;
       \State Compute the reconstruction loss $\mathcal{L}_{REC}$ via Eq.~\eqref{loss_rec};
       \State Compute the quality-aware contrastive loss $\mathcal{L}_{RCL}$ via Eq.~\eqref{loss_con};
       \State Construct the global consensus representation $H$ via Eq.~\eqref{eq_global_fusion};
       \State Compute the mutual-information alignment loss $\mathcal{L}_{MI}$ via Eq.~\eqref{eq_mi_loss};
       \State Update network parameters by minimizing Eq.~\eqref{loss_total_final};
   \EndFor
   \State Obtain the final cluster labels $\mathbf{y}$ by performing KMeans on the global consensus representation $H$.
   \State \Return $\mathbf{y}$
\end{algorithmic}
\end{algorithm}

\section{Experiments}

To comprehensively evaluate QARMVC, we conduct experiments from six aspects. First, we compare QARMVC with several representative deep multi-view clustering methods to assess its overall clustering performance. Second, we analyze the effectiveness of the proposed information bottleneck based quality estimation module by examining the correlation between the estimated scores and the actual contamination intensity. Third, we perform ablation studies to verify the contribution of each core component. Fourth, we visualize the learned latent representations and clustering structures to further illustrate the representation quality of QARMVC. Fifth, we investigate the sensitivity of QARMVC to key hyper-parameters, including the loss trade-off coefficients and the bottleneck dimension. Finally, we evaluate QARMVC in a comprehensive real-world scenario to examine its practical effectiveness under naturally existing quality variations.

\subsection{Experimental Settings}
\textbf{Datasets:} We conduct experiments on five widely used multi-view benchmark datasets. Scene-15~\cite{dai2013ensemble} comprises 4,485 images belonging to 15 classes, encompassing both indoor and outdoor environments. For each image, we extract GIST and PHOG features. MNIST-USPS~\cite{lecun1998gradient} consists of 5,000 handwritten digit images distributed across 10 digit categories. We utilize feature vectors from MNIST and USPS sources. LandUse-21~\cite{yang2010bag} contains 2,100 satellite imagery samples categorized into 21 classes. We employ PHOG and LBP features to represent the visual information. ALOI~\cite{geusebroek2005amsterdam} is a collection of 10,800 object images belonging to 100 categories. We adopt Color Similarity and HSV histograms. Finally, 100leaves~\cite{zheng2021collaborative} contains 1,600 samples from 100 categories. We extract FSM and SD features for analysis.

To simulate heterogeneous observation noise in real-world scenarios, we follow the noise injection protocol commonly used in existing robust multi-view clustering methods~\cite{dong2025rac,MVCAN}, and further extend it to model fine-grained variations in noise severity. Specifically, a proportion of samples is randomly selected for contamination according to $\eta \in \{10\%, 30\%, 50\%\}$. For each selected sample, Gaussian noise is injected with an intensity coefficient $\alpha \in \{0.2, 0.4, \dots, 1.0\}$. Accordingly, the contaminated view $\bar{x}$ is constructed as $\bar{x} = \alpha \cdot \delta + (1 - \alpha) \cdot x$, where $\delta$ denotes Gaussian noise.

\begin{table*}[!t]
    \centering
    \small
    \setlength{\tabcolsep}{4.5pt}
    \renewcommand{\arraystretch}{1.2}
    \caption{Clustering performance comparison on five benchmark datasets under varying heterogeneous observation noise ratios.}
    \label{tab_comp}
    \begin{tabular}{clccccccccccccccc}
    \toprule
    \multirow{2}*{Ratio} & \multirow{2}*{Methods} & \multicolumn{3}{c}{Scene15} & \multicolumn{3}{c}{MNIST-USPS} & \multicolumn{3}{c}{LandUse21} & \multicolumn{3}{c}{ALOI} & \multicolumn{3}{c}{100Leaves} \\
    \cmidrule(lr){3-5}\cmidrule(lr){6-8}\cmidrule(lr){9-11}\cmidrule(lr){12-14}\cmidrule(lr){15-17}
     & & ACC & NMI & ARI & ACC & NMI & ARI & ACC & NMI & ARI & ACC & NMI & ARI & ACC & NMI & ARI\\
    \midrule
    \multirow{8}*{10\%}
    & SURE      & 38.57 & 38.99 & 23.00 & \underline{90.98} & \underline{87.05} & \underline{82.21} & 24.19 & 25.86 & 9.54 & \textbf{59.26} & \underline{78.70} & \underline{43.28} & 42.75 & 76.07 & 24.20 \\
    & CANDY     & 38.10 & 36.96 & 21.87 & 75.86 & 74.76 & 66.15 & \underline{25.62} & \underline{30.03} & \underline{12.28} & 17.24 & 56.72 & 17.59 & 51.69 & 75.11 & 36.86 \\
    & DIVIDE    & 38.95 & 36.30 & 20.53 & 78.86 & 80.11 & 70.72 & 25.14 & 29.14 & 11.34 & 19.32 & 63.28 & 21.98 & 70.86 & \underline{86.42} & \underline{64.68} \\
    & MVCAN     & 26.91 & 27.24 & 14.09 & 56.30 & 51.87 & 38.75 & 20.33 & 23.77 & 7.46 & 40.83 & 64.77 & 31.09 & \underline{72.81} & 85.51 & 60.46 \\
    & RAC-DMVC   & \underline{40.32} & \underline{40.11} & \textbf{24.60} & 78.38 & 68.30 & 73.17 & 24.90 & 29.44 & \textbf{12.31} & 20.54 & 64.24 & 22.19 & 64.88 & 79.51 & 48.57 \\
    & MSDIB     & 34.22 & 34.48 & 19.32 & 61.08 & 65.72 & 51.13 & 23.85 & 28.84 & 10.66 & 45.46 & 69.99 & 33.18 & 65.63 & 80.84 & 50.52 \\
    & STCMC\_UR & 30.16 & 37.72 & 20.58 & 84.72 & 85.87 & 77.88 & 18.85 & 26.64 & 6.67 & 27.18 & 50.44 & 4.28 & 66.78 & 79.39 & 49.84 \\
    \midrule
    \rowcolor{blue!10} 
    & \textbf{QARMVC}& \textbf{43.83} & \textbf{40.32} & \underline{24.21} & \textbf{96.54} & \textbf{90.14} & \textbf{91.45} & \textbf{25.72} & \textbf{31.66} & 12.03 & \underline{49.05} & \textbf{78.75} & \textbf{49.65} & \textbf{77.44} & \textbf{89.33} & \textbf{69.45} \\
    \midrule
    \multirow{8}*{30\%}
    & SURE      & 35.67 & \underline{38.58} & \underline{21.19} & \underline{80.64} & 71.45 & 65.20 & 22.03 & 21.34 & 8.03 & \underline{41.56} & \underline{70.38} & \underline{34.46} & 44.19 & 71.63 & 27.44 \\
    & CANDY     & 32.40 & 28.20 & 15.96 & 66.50 & 72.24 & 61.86 & \underline{24.43} & 23.66 & 8.93 & 17.63 & 57.24 & 18.20 & 38.25 & 65.95 & 24.38 \\
    & DIVIDE    & \underline{38.19} & 35.51 & 19.54 & 78.76 & 67.85 & 58.58 & 24.29 & \underline{29.33} & \underline{10.79} & 17.89 & 58.59 & 18.80 & \underline{58.13} & \underline{76.91} & \underline{43.90} \\
    & MVCAN     & 26.73 & 24.30 & 13.02 & 49.86 & 46.87 & 32.64 & 15.14 & 16.19 & 4.21 & 12.75 & 30.59 & 6.45 & 54.31 & 74.74 & 38.28 \\
    & RAC-DMVC   & 30.26 & 28.08 & 13.94 & 70.62 & 53.29 & 59.86 & 22.38 & 22.68 & 8.30 & 17.23 & 58.12 & 18.70 & 50.06 & 70.78 & 33.21 \\
    & MSDIB     & 28.33 & 27.25 & 12.70 & 52.20 & 51.58 & 35.12 & 18.42 & 21.72 & 6.99 & 20.09 & 43.00 & 10.50 & 51.42 & 69.35 & 32.58 \\
    & STCMC\_UR & 27.67 & 34.33 & 15.09 & 79.26 & \underline{83.38} & \underline{74.60} & 17.95 & 25.51 & 6.03 & 18.50 & 42.78 & 3.02 & 50.15 & 69.75 & 32.73 \\
    \midrule
    \rowcolor{blue!10} 
    & \textbf{QARMVC}& \textbf{42.72} & \textbf{40.14} & \textbf{23.97} & \textbf{94.73} & \textbf{88.32} & \textbf{88.83} & \textbf{24.82} & \textbf{29.52} & \textbf{11.03} & \textbf{42.60} & \textbf{71.86} & \textbf{41.15} & \textbf{70.94} & \textbf{84.89} & \textbf{60.05} \\
    \midrule
    \multirow{8}*{50\%}
    & SURE      & \underline{30.42} & \underline{30.71} & \underline{15.81} & 64.92 & 58.82 & \underline{49.80} & 15.57 & 14.15 & 3.73 & \underline{20.50} & \underline{44.74} & 7.99 & 29.88 & 60.47 & 13.67 \\
    & CANDY     & 25.33 & 20.92 & 10.38 & 68.04 & 54.70 & 48.00 & 15.62 & 13.26 & 3.53 & 10.42 & 39.65 & 7.02 & 27.87 & 57.69 & 13.13 \\
    & DIVIDE    & 29.14 & 27.10 & 13.17 & \underline{73.28} & 60.61 & 48.59 & \underline{19.10} & \underline{21.02} & \underline{6.50} & 11.06 & 41.86 & 7.93 & \underline{43.31} & 67.46 & \underline{26.56} \\
    & MVCAN     & 19.60 & 14.92 & 7.64 & 47.02 & 44.27 & 26.83 & 12.13 & 12.03 & 3.98 & 5.99 & 16.17 & 1.35 & 41.94 & \underline{67.71} & 25.15 \\
    & RAC-DMVC   & 30.26 & 28.08 & 13.94 & 50.96 & 31.20 & 42.03 & 16.71 & 15.26 & 4.36 & 11.32 & 41.24 & \underline{8.02} & 38.31 & 64.17 & 23.08 \\
    & MSDIB     & 22.67 & 23.91 & 9.91 & 33.56 & 32.47 & 17.14 & 16.76 & 17.12 & 4.79 & 5.83 & 20.03 & 1.72 & 39.42 & 63.55 & 23.88 \\
    & STCMC\_UR & 25.68 & 25.19 & 9.53 & 56.42 & \underline{64.38} & 44.58 & 14.80 & 18.65 & 3.16 & 12.37 & 31.10 & 1.13 & 40.34 & 63.23 & 23.18 \\
    \midrule
    \rowcolor{blue!10} 
    & \textbf{QARMVC}& \textbf{35.42} & \textbf{33.63} & \textbf{19.12} & \textbf{93.74} & \textbf{85.69} & \textbf{86.33} & \textbf{22.45} & \textbf{24.84} & \textbf{9.02} & \textbf{23.12} & \textbf{50.02} & \textbf{16.75} & \textbf{54.19} & \textbf{75.35} & \textbf{38.62} \\
    \bottomrule
    \end{tabular}
    
    \vspace{2pt}
    \raggedright 
    \footnotesize 
    \hspace*{2em} 
    \textbf{Bold} indicates the best performance, and \underline{underline} indicates the second best.
\end{table*}

\noindent\textbf{Comparison Algorithms:}
To showcase the broad applicability and superior performance of QARMVC, we compare it against several state-of-the-art deep multi-view clustering methods, including noise-resilient methods designed to handle corrupted data (CANDY~\cite{guo2024robust}, DIVIDE~\cite{DIVIDE}, RAC-DMVC~\cite{dong2025rac}, and MVCAN~\cite{MVCAN}) and classical methods (MSDIB~\cite{hu2025multi}, STCMC\_UR~\cite{hu2025self} and SURE~\cite{yang2022robust}). For fair comparison, all baselines are reproduced using their official implementations.

\noindent\textbf{Implementation Details.}
In this study, all experiments are implemented using the PyTorch framework~\cite{pytorch} on a workstation equipped with a single NVIDIA RTX 4090 GPU. The proposed QARMVC is trained in an end-to-end manner using the Adam optimizer~\cite{adam}, with the learning rate fixed at $1 \times 10^{-3}$. The total training process spans $200$ epochs with a batch size of $128$. As for the hyperparameters in the overall objective function Eq.~\eqref{loss_total_final}, the trade-off weights are consistently set as $\lambda_1=0.1$, $\lambda_2=0.1$, and $\lambda_3=0.1$.

\subsection{Comparison Experiments}

In this subsection, we compare QARMVC with several representative baselines under different heterogeneous observation noise ratios. The quantitative results are reported in Table~\ref{tab_comp}, where the best and second-best results are highlighted in bold and underlined, respectively. Overall, QARMVC achieves the best or highly competitive performance on most dataset-metric pairs, demonstrating its effectiveness in noisy multi-view clustering.

As the noise ratio increases from 10\% to 50\%, the performance of most baselines degrades substantially, whereas QARMVC remains much more stable. This result suggests that the proposed quality-aware framework can effectively suppress the influence of unreliable observations and preserve more discriminative clustering structures under severe heterogeneous noise.

In particular, on MNIST-USPS, QARMVC consistently achieves the best results across all three noise settings, reaching 96.54/90.14/91.45, 94.73/88.32/88.83, and 93.74/85.69/86.33 in terms of ACC/NMI/ARI, respectively. Under 50\% noise, it improves ACC over the strongest baseline by 20.46 percentage points, which clearly verifies its robustness in heavily corrupted scenarios. Similar improvements can also be observed on Scene15 and ALOI, especially under medium and high noise ratios.

Although QARMVC is not always ranked first on a few relatively easy cases, it shows a clear advantage as the noise level becomes higher. In summary, these results demonstrate that QARMVC not only yields strong clustering performance, but also exhibits superior robustness against heterogeneous observation noise.

\begin{figure}[t]
  \centering

  \begin{subfigure}{0.48\linewidth}
    \centering
    \includegraphics[width=\linewidth]{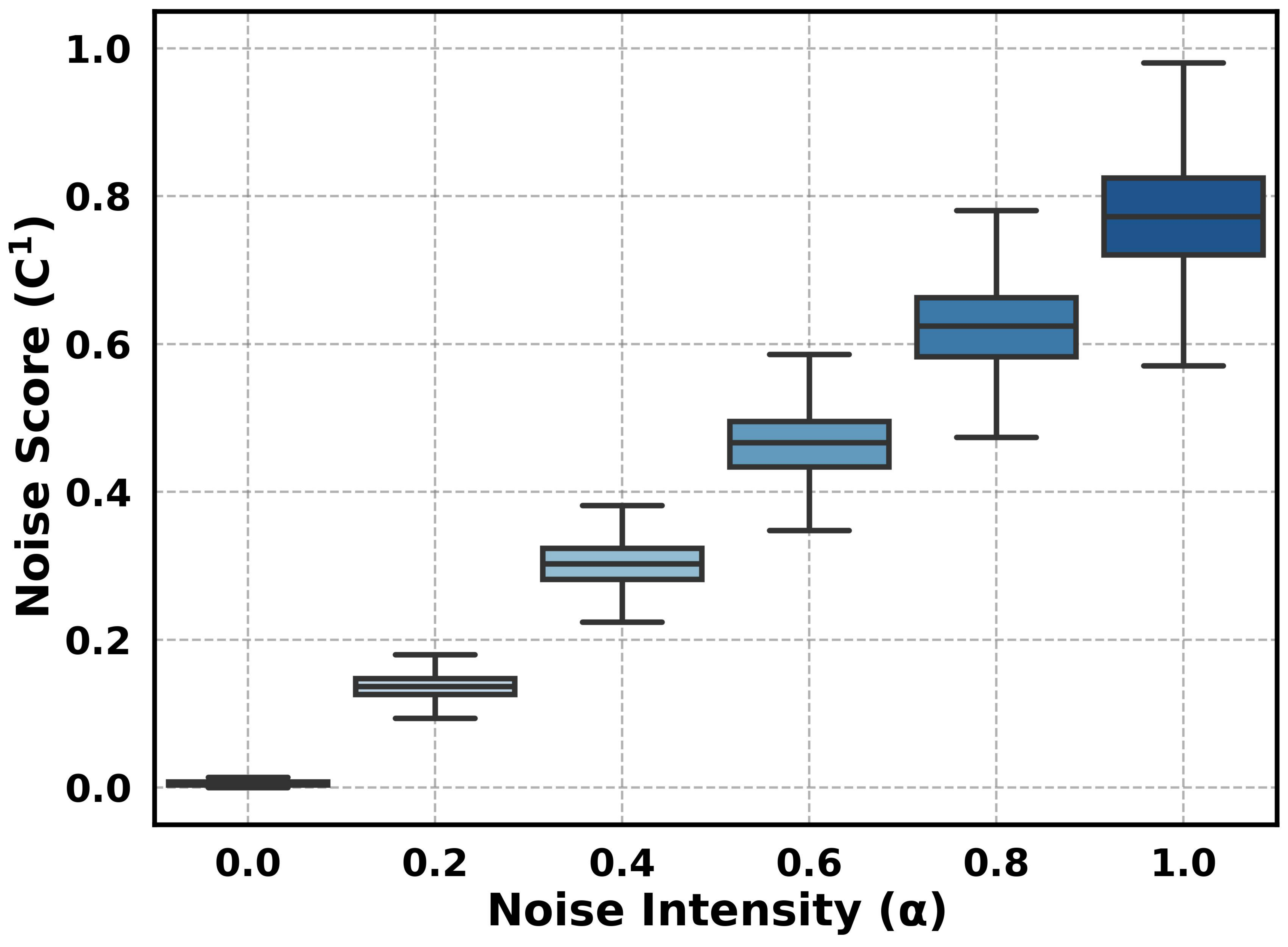}
    \caption{ALOI: Color Similarity}
    \label{fig:aloi_v1}
  \end{subfigure}
  \hfill
  \begin{subfigure}{0.48\linewidth}
    \centering
    \includegraphics[width=\linewidth]{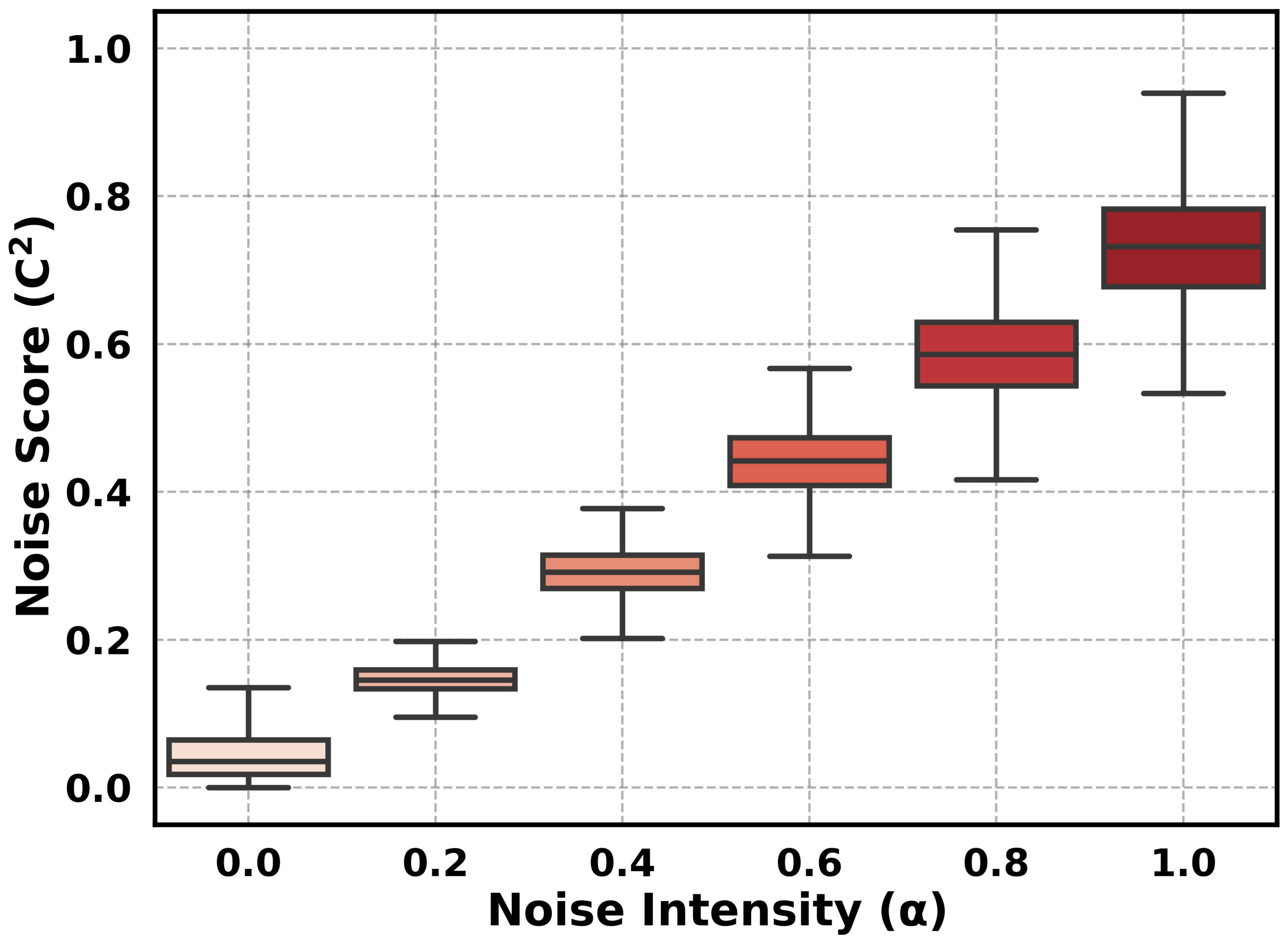}
    \caption{ALOI: HSV Histogram}
    \label{fig:aloi_v2}
  \end{subfigure}

  \vspace{2mm}

  \begin{subfigure}{0.48\linewidth}
    \centering
    \includegraphics[width=\linewidth]{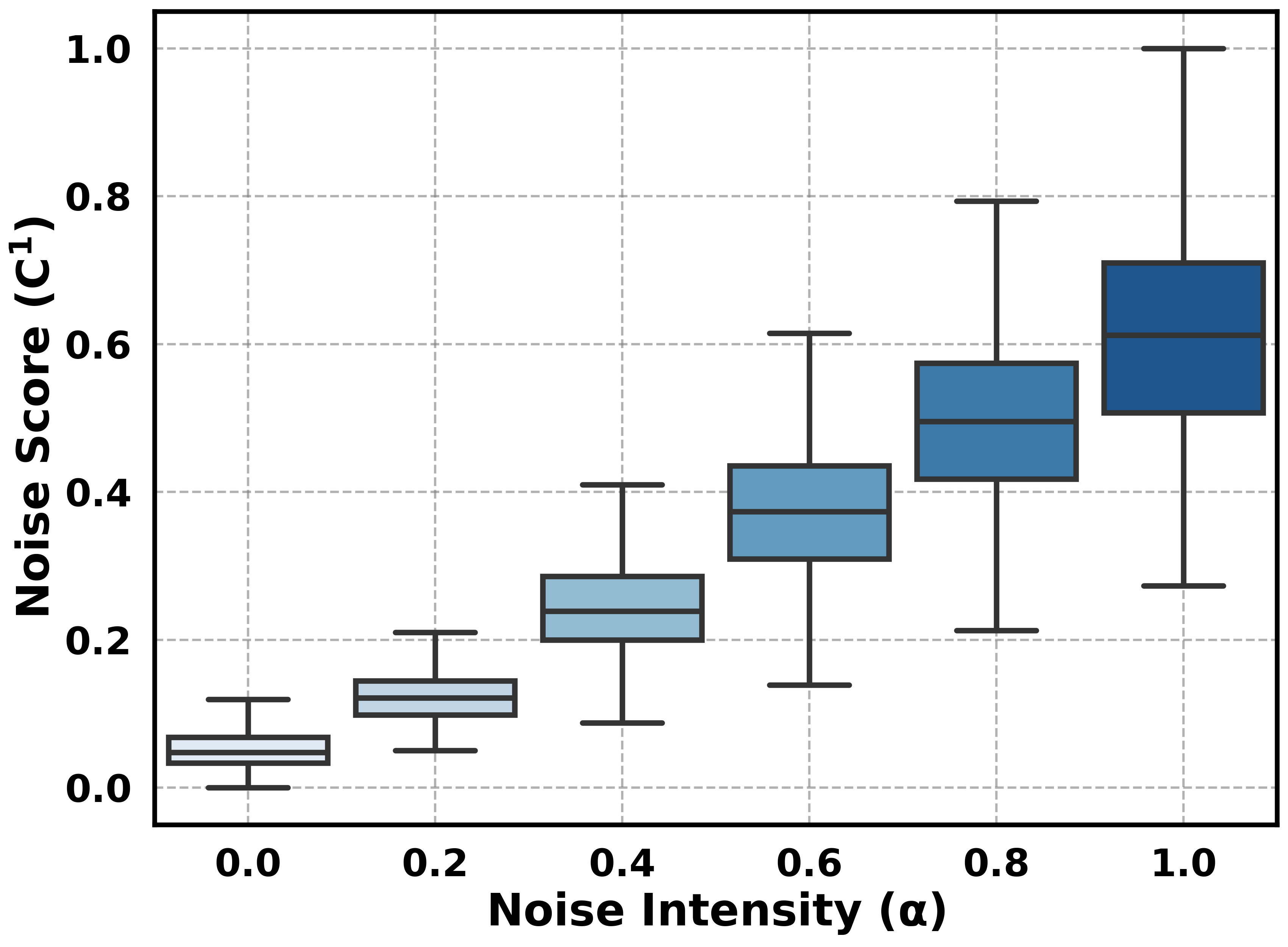}
    \caption{Scene15: GIST}
    \label{fig:scene15_v1}
  \end{subfigure}
  \hfill
  \begin{subfigure}{0.48\linewidth}
    \centering
    \includegraphics[width=\linewidth]{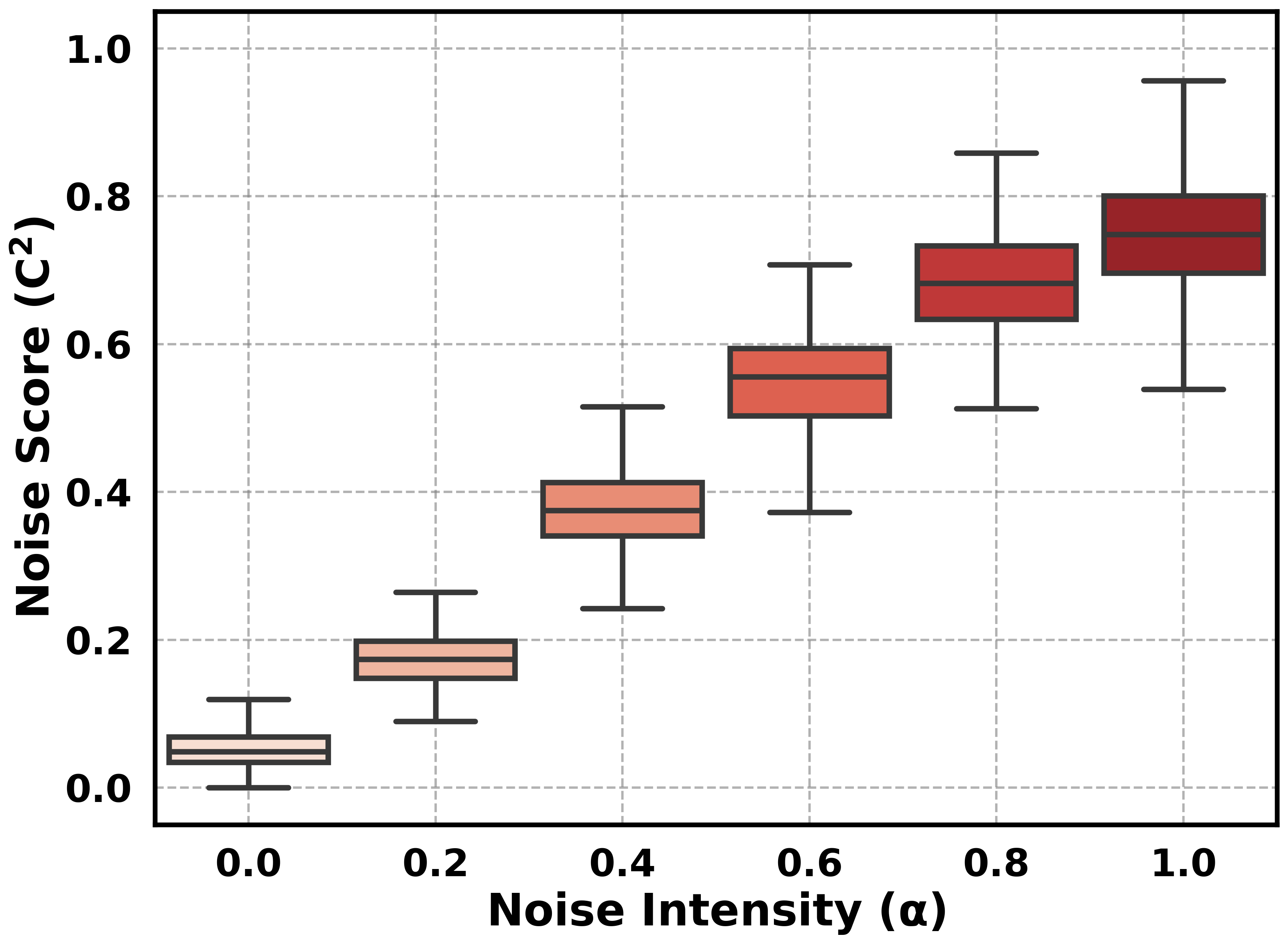}
    \caption{Scene15: PHOG}
    \label{fig:scene15_v2}
  \end{subfigure}

  \caption{Noise score analysis on the ALOI and Scene15 datasets.}
  \label{fig:quality_score_analysis}
  \vspace{-3mm}
\end{figure}

\begin{table}[t]
    \centering
    \vspace{-2mm} 
    \caption{Analysis between noise scores($C^v$) and intensities($\alpha$).}
    \label{tab:correlation_coef}
    
    \renewcommand{\arraystretch}{0.6} 
    \setlength{\tabcolsep}{2.7pt} 
    
    \small
    \begin{tabular}{llccc}
    \toprule
    \textbf{Dataset} & \textbf{Feature} & \textbf{BottleneckDim} & \textbf{Pearson} & \textbf{Spearman} \\
    \midrule
    \multirow{2}{*}{Scene15} 
         & GIST & 7  & 0.933 & 0.904 \\
         & PHOG & 20 & 0.979 & 0.925 \\
    \midrule
    \multirow{2}{*}{MNIST-USPS} 
         & MNIST & 520 & 0.921 & 0.779 \\
         & USPS  & 100 & 0.927 & 0.824 \\
    \midrule
    \multirow{2}{*}{LandUse-21} 
         & PHOG & 7  & 0.844 & 0.834 \\
         & LBP  & 20 & 0.977 & 0.920 \\
    \midrule
    \multirow{2}{*}{ALOI} 
         & CS & 20 & 0.991 & 0.932 \\
         & HSV & 20 & 0.979 & 0.921 \\
    \midrule
    \multirow{2}{*}{100Leaves} 
         & FSM & 32 & 0.943 & 0.882 \\
         & SD & 32 & 0.917 & 0.822 \\
    \bottomrule
    \end{tabular}
    \vspace{-3mm} 
\end{table}

\begin{figure*}[t]
\centering
\small
\begin{minipage}{0.24\linewidth}
    \centerline{\includegraphics[width=\textwidth]{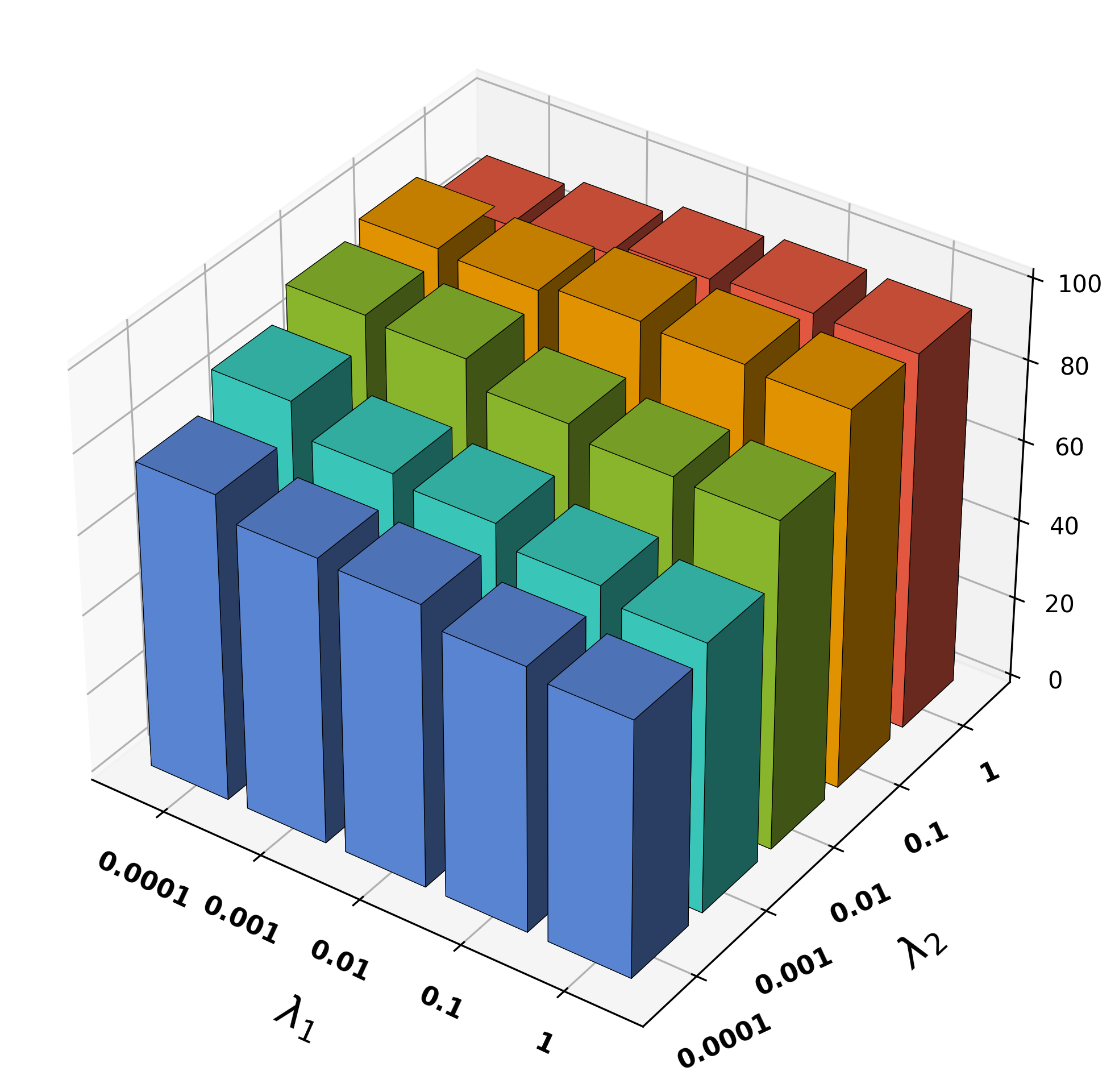}}
    \centerline{ACC vs. $\lambda_1$ and $\lambda_2$}
\end{minipage}
\hfill 
\begin{minipage}{0.24\linewidth}
    \centerline{\includegraphics[width=\textwidth]{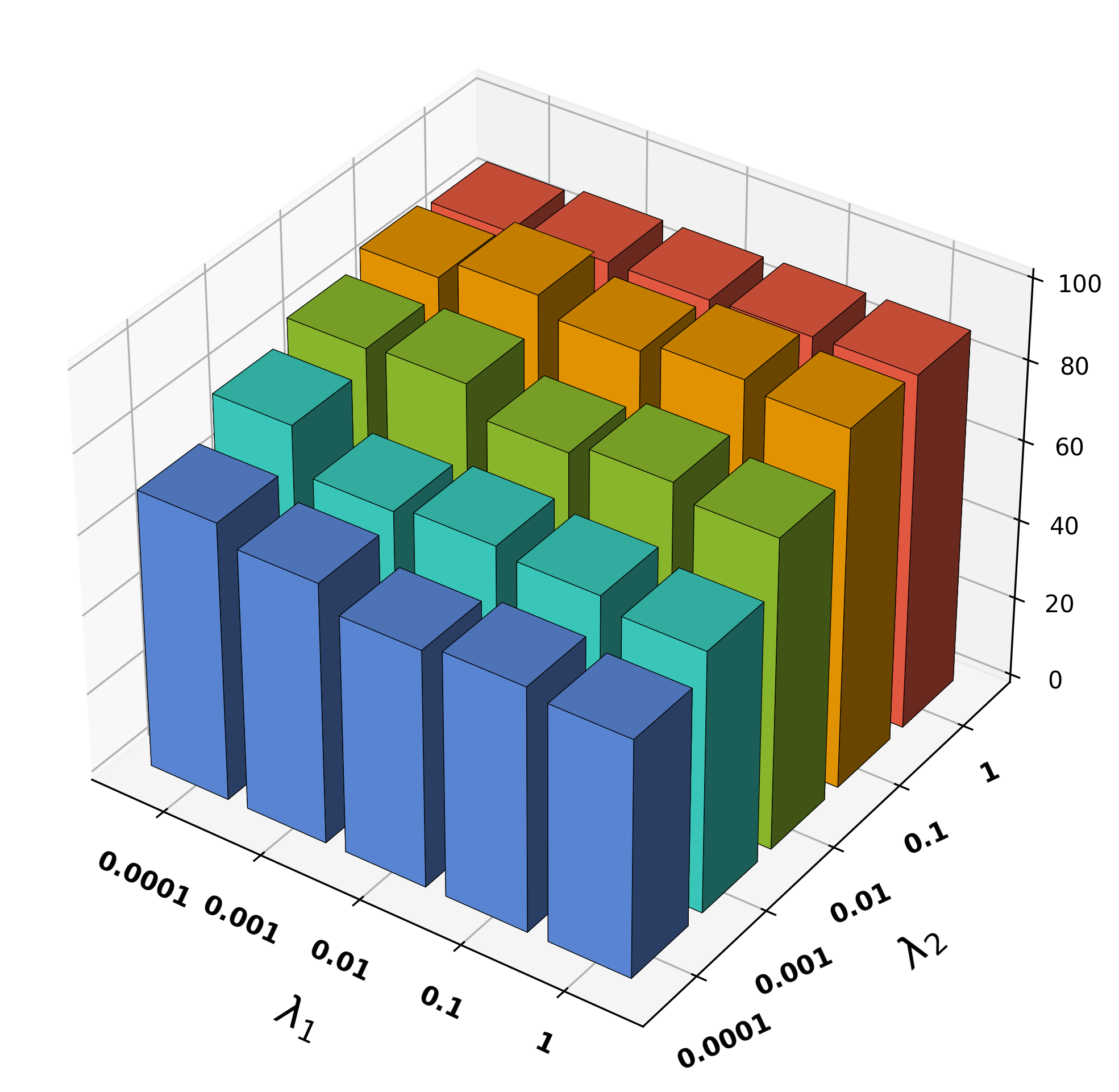}}
    \centerline{MMI vs. $\lambda_1$ and $\lambda_2$}
\end{minipage}
\hfill
\begin{minipage}{0.24\linewidth}
    \centerline{\includegraphics[width=\textwidth]{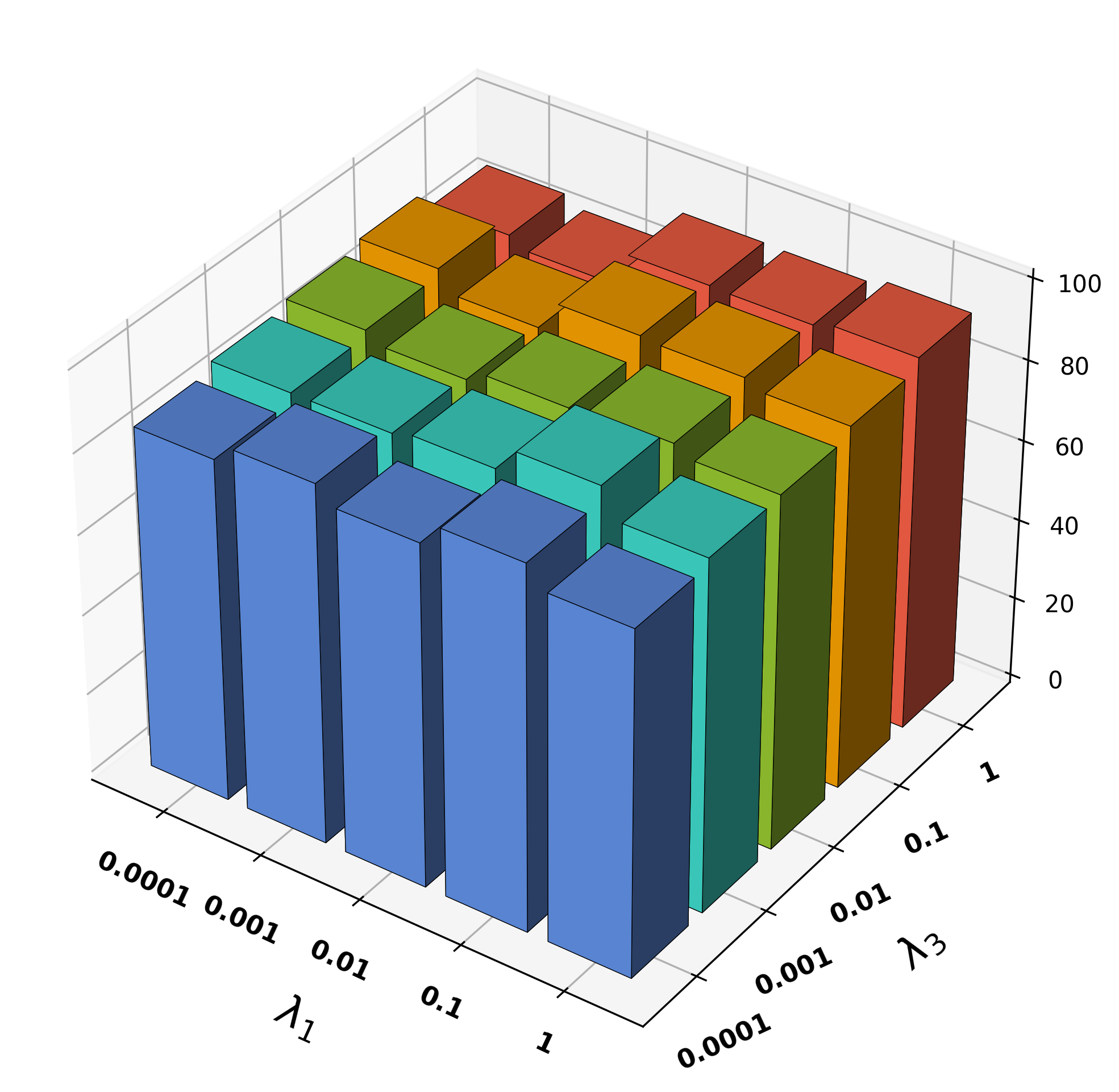}}
    \centerline{ACC vs. $\lambda_1$ and $\lambda_3$}
\end{minipage}
\hfill
\begin{minipage}{0.24\linewidth}
    \centerline{\includegraphics[width=\textwidth]{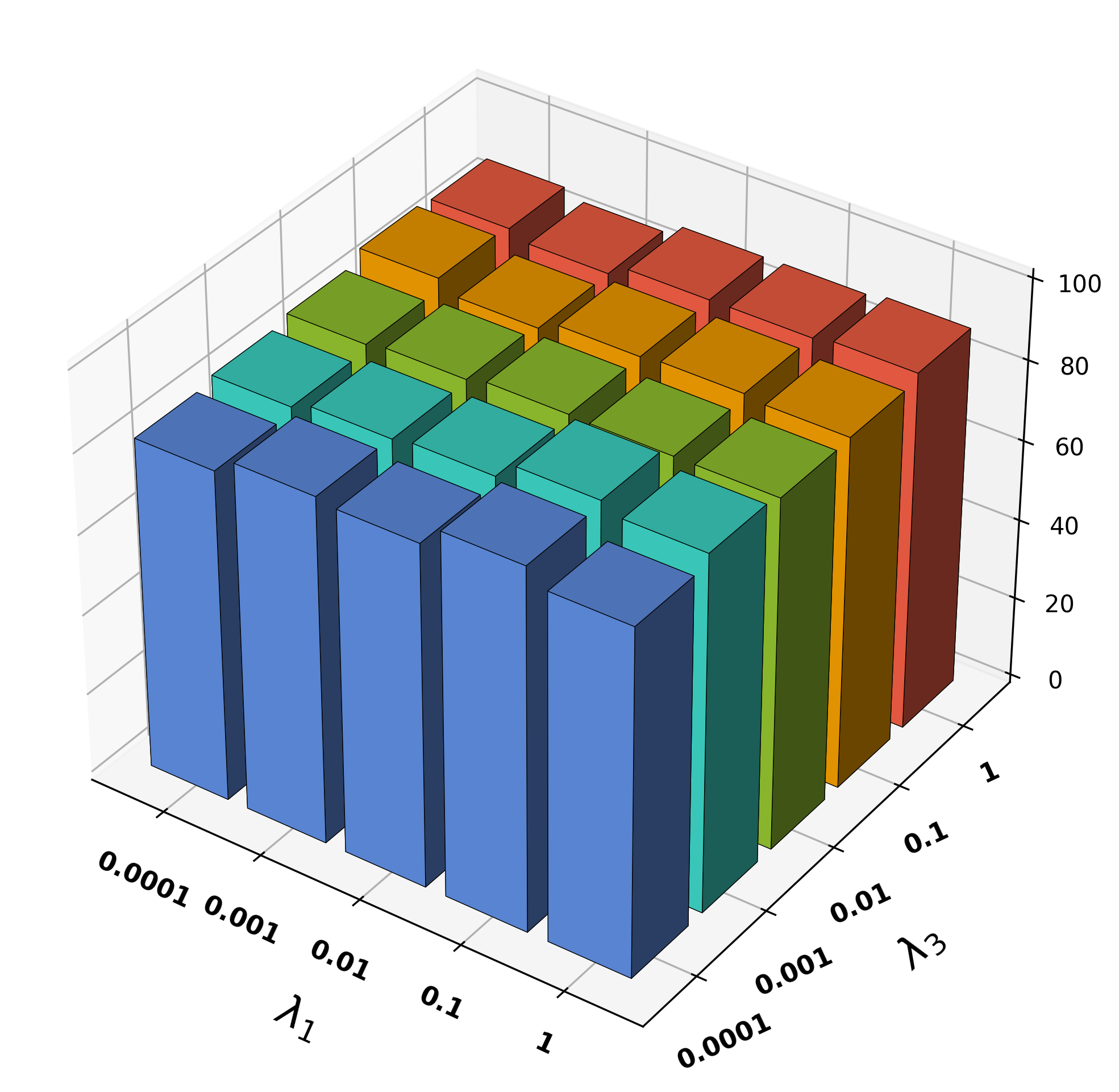}}
    \centerline{NMI vs. $\lambda_1$ and $\lambda_3$}
\end{minipage}
\caption{Sensitivity analysis on MNIST-USPS dataset with 10\% Noise.}
\label{fig:sen_analysis_4in1}
\end{figure*}

\begin{figure*}
\centering
\begin{minipage}{0.22\linewidth}
\centerline{\includegraphics[width=1\textwidth]{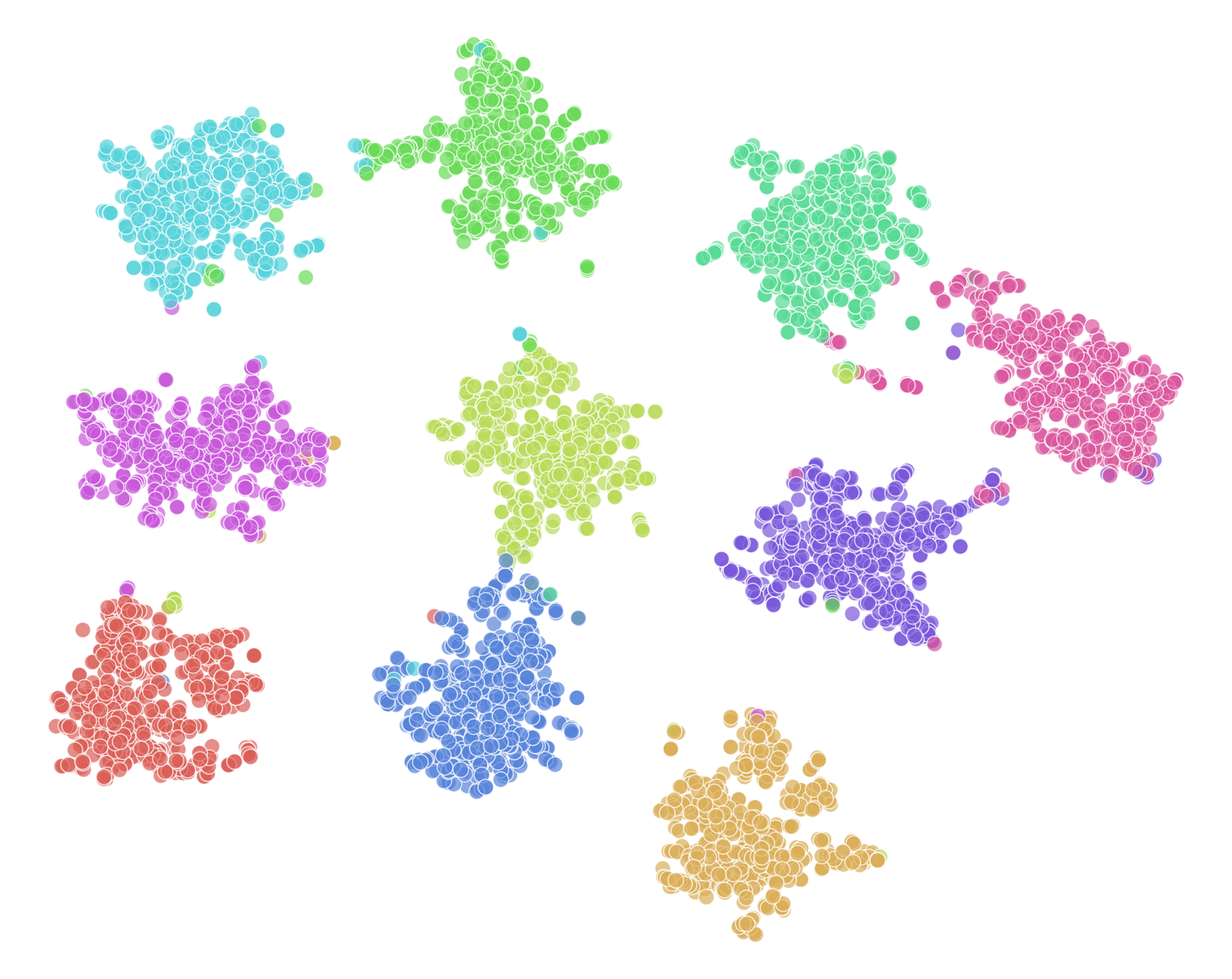}}
\vspace{5pt}
\centerline{{(a) Ours}}
\end{minipage}\hspace{5pt}
\begin{minipage}{0.22\linewidth}
\centerline{\includegraphics[width=1\textwidth]{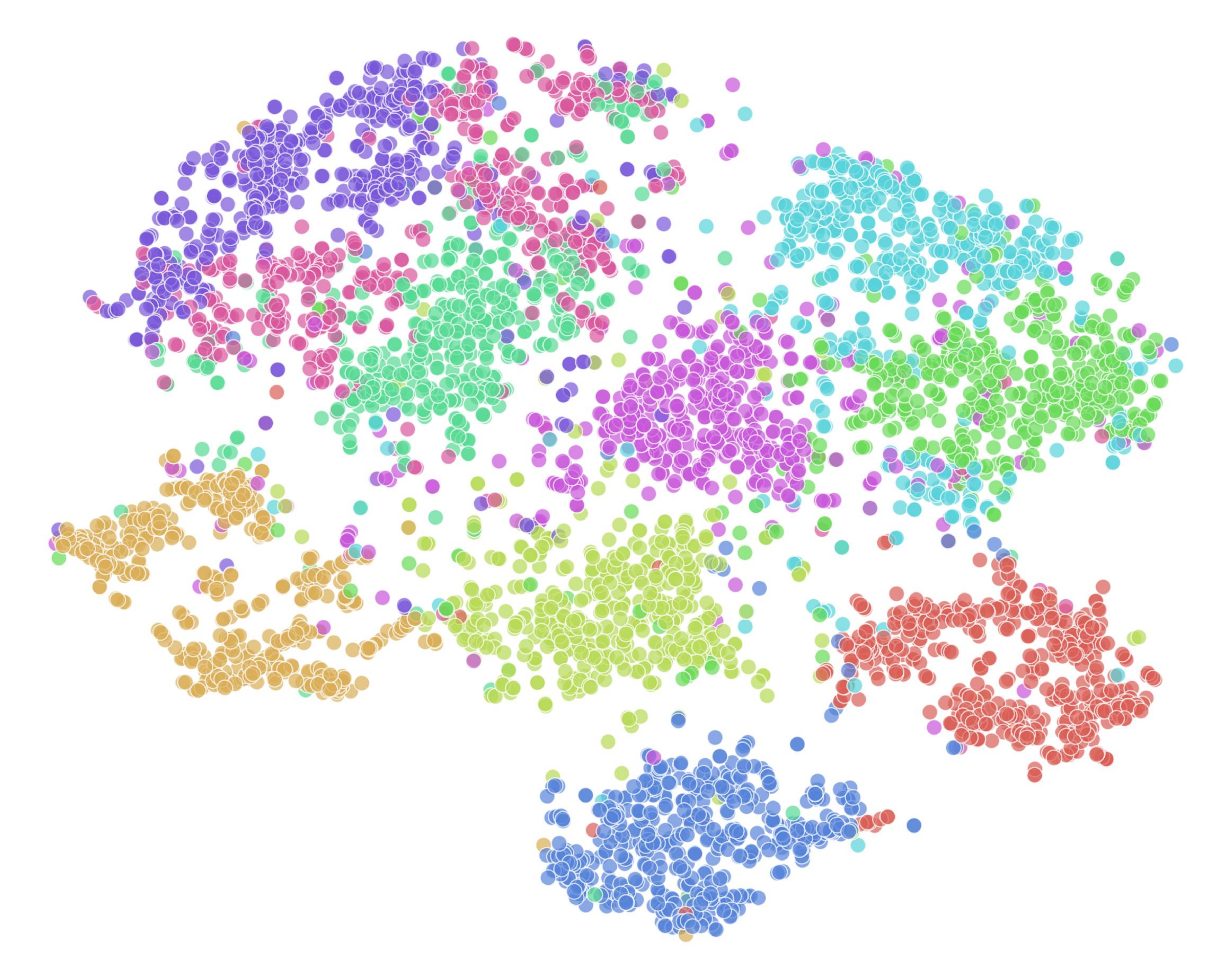}}
\vspace{5pt}
\centerline{{(b) DIVIDE}}
\end{minipage}\hspace{5pt}
\begin{minipage}{0.22\linewidth}
\centerline{\includegraphics[width=1\textwidth]{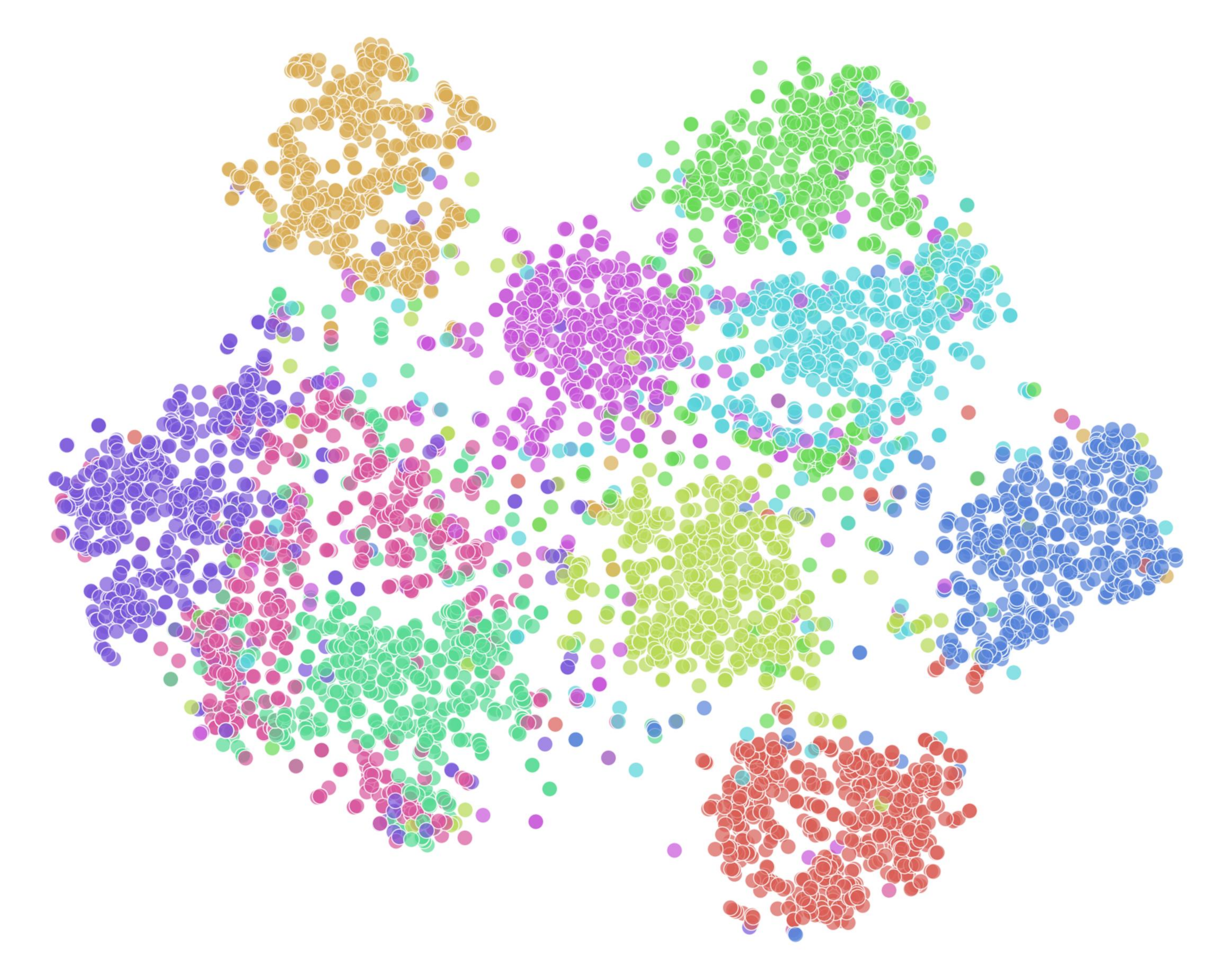}}
\vspace{5pt}
\centerline{{(c) RAC-DMVC}}
\end{minipage}\hspace{5pt}
\begin{minipage}{0.22\linewidth}
\centerline{\includegraphics[width=1\textwidth]{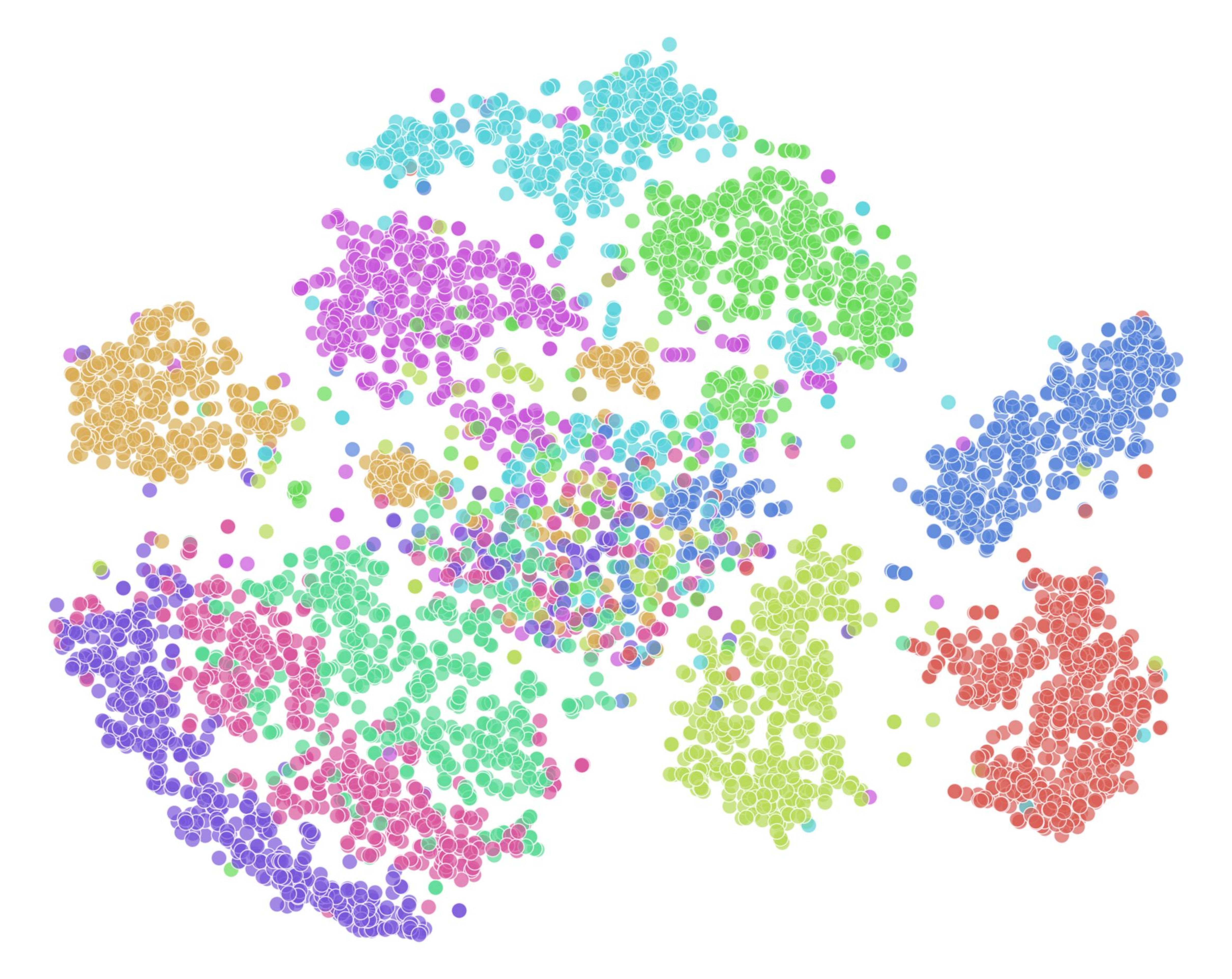}}
\vspace{5pt}
\centerline{{(d) MVCAN}}
\end{minipage}
\caption{Visualization on MNIST-USPS dataset with 10\% Noise.}
\label{t_sne}
\vspace{-10pt}
\end{figure*}

\subsection{Quality Score Analysis}
\label{sec:quality_analysis}
To validate the capability of QARMVC in accurately perceiving contamination intensity, we conduct a quantitative analysis under the 50\% noise ratio setting. Specifically, Figure~\ref{fig:quality_score_analysis} presents the distributions of the estimated noise scores ($C^v$) with respect to different noise intensities ($\alpha$) on the ALOI and Scene15 datasets. As the noise intensity increases, the estimated scores in all views exhibit a clear upward trend, and the corresponding box plots show a strong monotonic relationship between the estimated contamination level and the actual corruption intensity. Moreover, no disproportionately extreme outliers are observed in the estimated score distributions, suggesting that the proposed estimator retains sufficient resolution across different noise levels. This observation is further supported by the quantitative results in Table~\ref{tab:correlation_coef}, where consistently high Pearson and Spearman correlation coefficients are achieved across different datasets and feature views. These results confirm that the proposed estimator can effectively capture contamination intensity, making $Q^v=(1-C^v)^2$ a reliable measure of data quality for quality-aware learning and fusion.

\subsection{Ablation Study}
Table~\ref{tab:ablation_results} reports the ablation results on MNIST-USPS and 100leaves under 50\% noise. The full QARMVC achieves the best performance on both datasets, demonstrating the effectiveness of jointly optimizing reconstruction, quality-aware contrastive learning, and cross-view alignment. Among these components, the quality-aware contrastive objective is the most critical, as removing $\mathcal{L}_{RCL}$ causes a dramatic performance drop on both datasets. Removing $\mathcal{L}_{REC}$ or $\mathcal{L}_{MI}$ also consistently degrades the results, indicating that preserving view-specific information and enforcing semantic consistency are both necessary for robust fusion. In addition, QARMVC consistently outperforms Standard CL, which verifies the advantage of the proposed quality-aware contrastive strategy over indiscriminate pairwise alignment. The much weaker performance of the double-ablation variants further suggests that no single objective is sufficient, and the superiority of the full model comes from the cooperation of all components.

\begin{table}
  \centering
  \caption{Ablation study under 50\% noise ratio.}
  \label{tab:ablation_results}
  \setlength{\tabcolsep}{4pt}
  \small
  \begin{tabular}{l ccc ccc}
    \toprule
    \multirow{2}{*}{\textbf{Settings}} & \multicolumn{3}{c}{\textbf{MNIST-USPS}} & \multicolumn{3}{c}{\textbf{100leaves}} \\
    \cmidrule(lr){2-4} \cmidrule(lr){5-7}
    & \textbf{ACC} & \textbf{NMI} & \textbf{ARI} & \textbf{ACC} & \textbf{NMI} & \textbf{ARI} \\
    \midrule
    w/o $\mathcal{L}_{RCL}$ \& $\mathcal{L}_{MI}$ & 28.32 & 25.45 & 12.88 & 40.34 & 73.55 & 24.01 \\
    w/o $\mathcal{L}_{REC}$ \& $\mathcal{L}_{MI}$ & 78.23 & 70.44 & 68.55 & 52.50 & 74.24 & 37.23 \\
    w/o $\mathcal{L}_{REC}$ \& $\mathcal{L}_{RCL}$ & 28.78 & 26.56 & 12.22 & 40.13 & 73.13 & 23.54 \\
    w/o $\mathcal{L}_{MI}$                         & 88.85 & 82.64 & 79.77 & 51.33 & 73.53 & 35.23 \\
    w/o $\mathcal{L}_{RCL}$                         & 30.12 & 27.33 & 13.58 & 41.81 & 68.47 & 24.89 \\
    w/o $\mathcal{L}_{REC}$                         & 80.54 & 73.58 & 69.11 & 52.38 & 74.65 & 37.78 \\
    Standard CL                                     & 88.23 & 81.34 & 78.22 & 50.19 & 73.30 & 34.22 \\
    \midrule
    \textbf{QARMVC (Full)}                          & \textbf{93.74} & \textbf{85.69} & \textbf{86.33} & \textbf{54.19} & \textbf{75.35} & \textbf{38.62} \\
    \bottomrule
  \end{tabular}

    \vspace{2pt}
    \raggedright 
    \footnotesize 
    \hspace*{2em} 
    \textbf{Bold} indicates the best performance,
\end{table}

\subsection{Parameter Analysis}
\textbf{Loss function hyperparameter analysis.} As shown in Figure~\ref{fig:sen_analysis_4in1}, QARMVC is relatively robust to the choices of $\lambda_1$, $\lambda_2$, and $\lambda_3$ on MNIST-USPS with 10\% noise. With respect to the interaction between $\lambda_1$ and $\lambda_2$, better results are generally obtained when $\lambda_2$ is set to a moderate value, while overly small $\lambda_2$ leads to clear performance degradation, indicating that the quality-aware contrastive objective should maintain sufficient influence during training. A similar trend can also be observed for the combination of $\lambda_1$ and $\lambda_3$, where the model achieves stronger performance in the middle parameter region, whereas extreme settings tend to weaken clustering quality. Overall, the best results are obtained by balanced parameter combinations rather than boundary values, which suggests that QARMVC does not rely on delicate hyperparameter tuning and maintains stable performance over a reasonably wide range.

\begin{figure}[t]
  \centering
  \begin{subfigure}{0.48\linewidth}
    \centering
    \includegraphics[width=\linewidth]{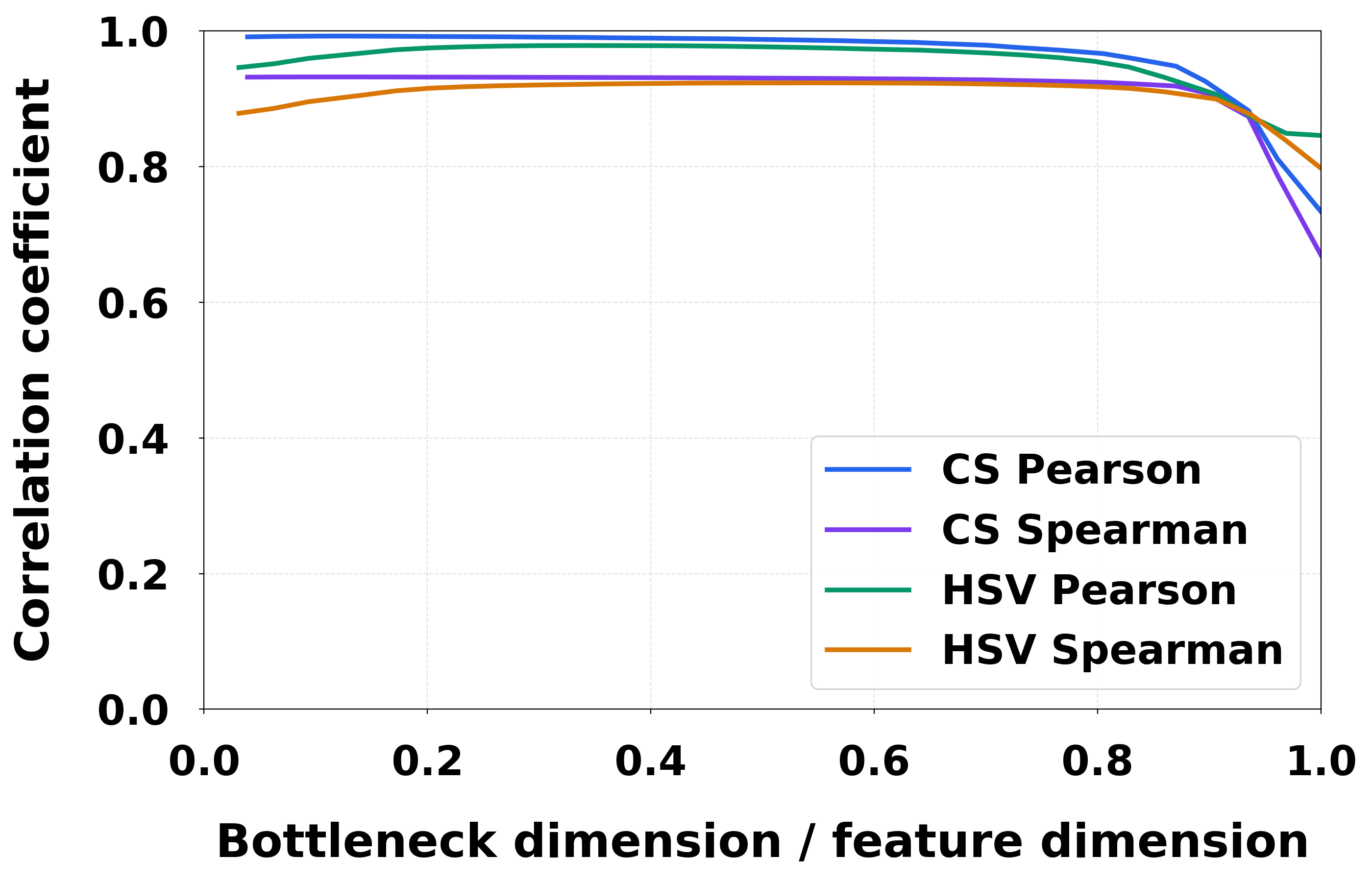}
    \caption{ALOI}
    \label{fig:aloi_bottledim}
  \end{subfigure}
  \hfill
  \begin{subfigure}{0.48\linewidth}
    \centering
    \includegraphics[width=\linewidth]{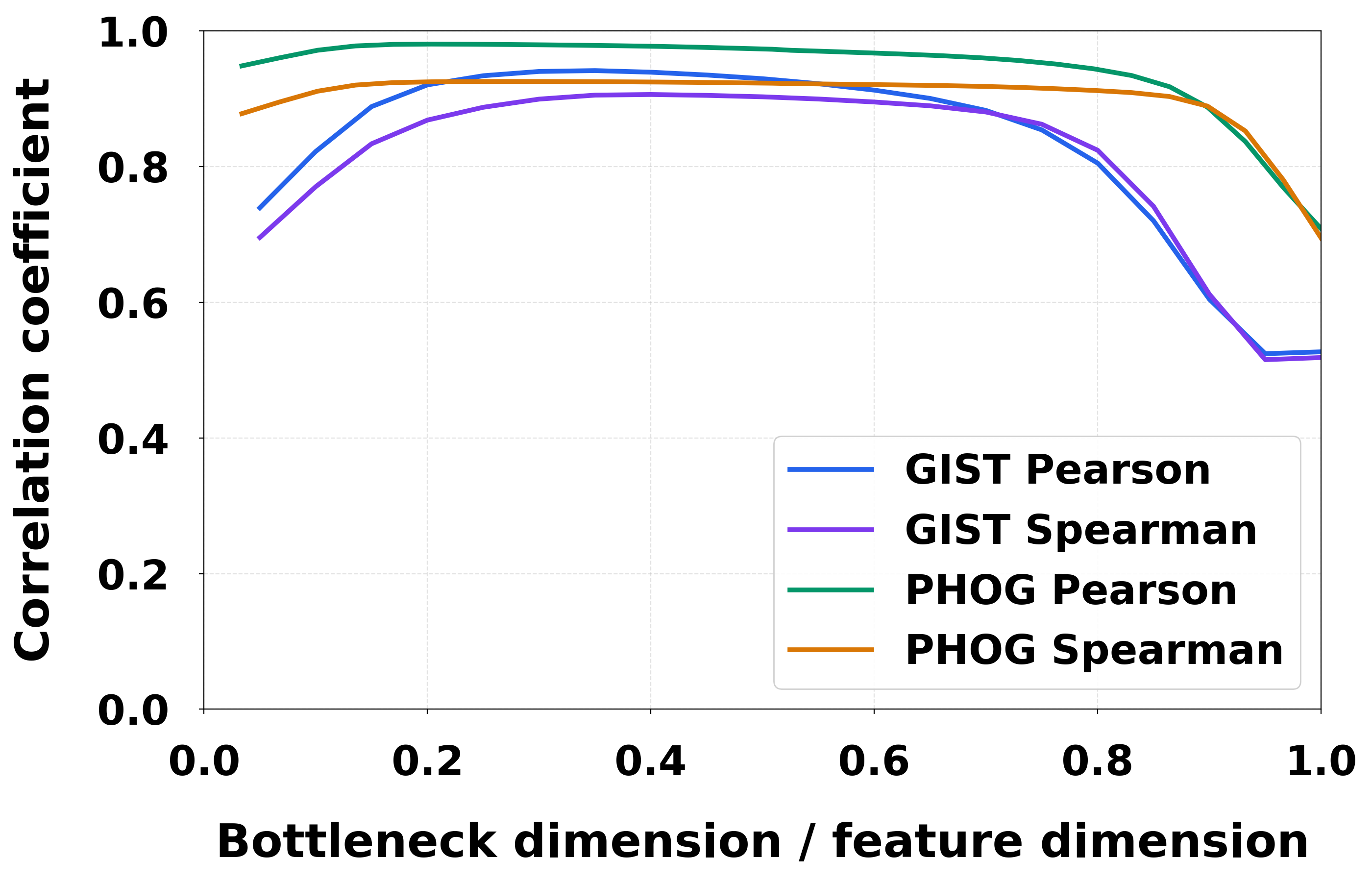}
    \caption{Scene15}
    \label{fig:scene15_bottledim}
  \end{subfigure}
  \caption{Correlation analysis of estimated noise scores under different bottleneck dimensions.}
  \label{fig:bottleneck_analysis}
  \vspace{-3mm}
\end{figure}

\noindent\textbf{Bottleneck dimension analysis.} We further study the effect of the bottleneck dimension on the quality estimation module in Figure~\ref{fig:bottleneck_analysis}. On both ALOI and Scene15, the Pearson and Spearman correlation coefficients remain consistently high across a broad range of bottleneck ratios, indicating that the proposed estimator is not overly sensitive to this hyperparameter and thus enjoys satisfactory robustness. Meanwhile, the best or near-best performance is generally achieved in the middle region. In contrast, overly small bottleneck dimensions may lead to insufficient information preservation, while excessively large ones weaken the desired compactness of the latent representation. Therefore, we recommend setting the bottleneck dimension to approximately $1/2$ of the input feature dimension in practice.

\subsection{Visualization Analysis}
To intuitively evaluate the learned representations, we visualize the latent embedding spaces of QARMVC and representative baselines on the MNIST-USPS dataset (10\% noise) via $t$-SNE~\cite{TSNE} in Figure~\ref{t_sne}. While baseline methods exhibit blurred boundaries and considerable overlap indicating vulnerability to noise, QARMVC generates a highly discriminative latent space characterized by high intra-cluster compactness and clear inter-cluster separability. This distinct structure confirms that our quality-aware framework effectively filters out noise interference to yield robust and consistent semantics, significantly surpassing competing methods.

\subsection{Application to Comprehensive Scenario}
We further evaluate QARMVC on the SUNRGBD dataset~\cite{song2015sun} without following the noise injection protocol, in order to verify its effectiveness in a comprehensive real-world scenario.

SUNRGBD is a representative RGB-D scene understanding dataset collected from real indoor environments. With multi-modal observations and substantial real-world variations, such as illumination changes, viewpoint shifts, background clutter, object occlusion, and sensor-induced quality fluctuations, it provides a challenging benchmark for robust multi-view clustering. We adopt the two-view setting and directly evaluate all methods on the original data.

\begin{figure}[t]
    \centering
    \includegraphics[width=\linewidth]{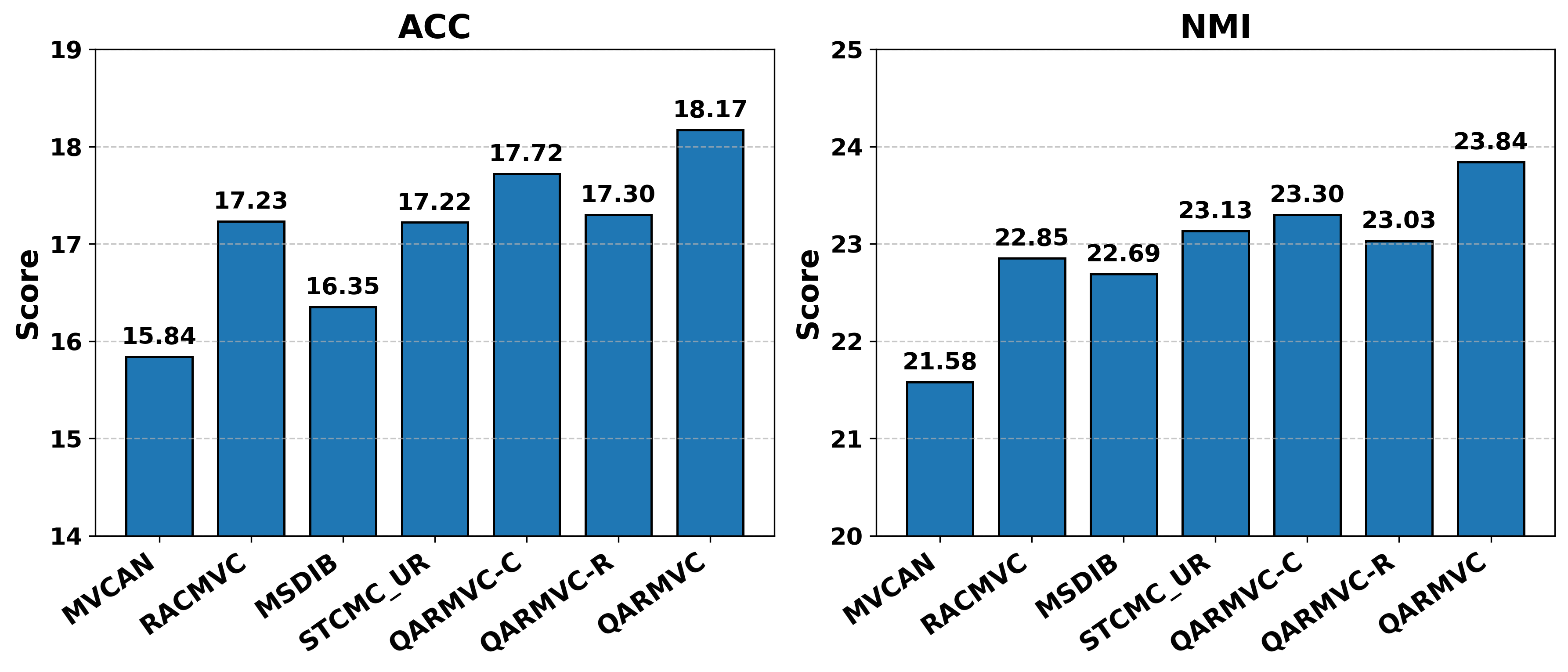}
    \caption{ACC and NMI comparison on the SUNRGBD dataset. QARMVC-C and QARMVC-R denote the variants of QARMVC using constant and random scores, respectively.}
    \label{fig:sunrgbd_realworld}
\end{figure}

As shown in Fig.~\ref{fig:sunrgbd_realworld}, QARMVC achieves the best ACC and NMI, reaching 18.17 and 24.49, respectively. It outperforms the strongest baseline by 0.94 points in ACC and 1.36 points in NMI, which suggests its effectiveness under naturally existing quality variations.

When the learned quality score is replaced by a constant or random one, the performance consistently drops to 17.72/23.30 and 17.30/23.03 in terms of ACC/NMI, respectively. This confirms that the improvement is brought by informative quality estimation rather than naive weighting.
These results provide preliminary evidence of the practical applicability of QARMVC on real-world multi-view data.

\section{Conclusion}
In this paper, we identify and address the challenge of heterogeneous observation noise in multi-view clustering. To this end, we propose Quality-Aware Robust Multi-View Clustering (QARMVC), a novel framework that leverages an information bottleneck mechanism to quantify instance-level data quality. These quality scores guide a weighted contrastive objective and a global-local alignment module, enabling the model to suppress noisy anchors and align distorted views through a robust global consensus. Extensive experiments on multiple benchmarks demonstrate the effectiveness of QARMVC, especially under varying noise intensities. Nevertheless, the current framework is mainly designed for unstructured random noise and may be less effective for more structured noise patterns. Exploring more adaptive robust multi-view clustering under broader real-world noise settings will be an important direction for future work.

\bibliographystyle{ACM-Reference-Format}
\bibliography{sample-base}

\end{document}